\documentclass[sigconf]{acmart}

\AtBeginDocument{%
  }

\copyrightyear{2023}
\acmYear{2023}
\setcopyright{acmlicensed}\acmConference[MM '23]{Proceedings of the 31st ACM International Conference on Multimedia}{October 29-November 3, 2023}{Ottawa, ON, Canada}
\acmBooktitle{Proceedings of the 31st ACM International Conference on Multimedia (MM '23), October 29-November 3, 2023, Ottawa, ON, Canada}
\acmPrice{15.00}
\acmDOI{10.1145/3581783.3612100}
\acmISBN{979-8-4007-0108-5/23/10}

\usepackage{multirow}
\usepackage{graphics}
\usepackage{bbding}
\usepackage{balance}

\begin{document}

\title{Implicit Obstacle Map-driven Indoor Navigation Model for Robust Obstacle Avoidance}


\author{Wei Xie}
\authornotemark[2]
\affiliation{%
  \institution{Nanjing University of Science and Technology}
  \city{Nanjing}
  \country{China}}
\email{xieweiayy@njust.edu.cn}

\author{Haobo Jiang}
\authornotemark[2]
\affiliation{%
  \institution{Nanjing University of Science and Technology}
  \city{Nanjing}
  \country{China}}
\email{jiang.hao.bo@njust.edu.cn}

\author{Shuo Gu}
\authornotemark[2]
\affiliation{%
  \institution{Nanjing University of Science and Technology}
  \city{Nanjing}
  \country{China}}
\email{shuogu@njust.edu.cn}

\author{Jin Xie}
\authornote{Corresponding author.}
\authornote{PCA Lab, Key Lab of Intelligent Perception and Systems for High-Dimensional Information of Ministry of Education, Jiangsu Key Lab of Image and Video Understanding for Social Security, School of Computer Science and Engineering, Nanjing University of Science and Technology.}
\affiliation{%
  \institution{Nanjing University of Science and Technology}
  \city{Nanjing}
  \country{China}}
\email{csjxie@njust.edu.cn}






\renewcommand{\shortauthors}{Wei Xie, Haobo Jiang, Shuo Gu, \& Jin Xie}

\begin{abstract}
Robust obstacle avoidance is one of the critical steps for successful goal-driven indoor navigation tasks. 
Due to the obstacle missing in the visual image and the possible missed detection issue, visual image-based obstacle avoidance techniques still suffer from unsatisfactory robustness. 
To mitigate it, in this paper, we propose a novel implicit obstacle map-driven indoor navigation framework for robust obstacle avoidance, where an implicit obstacle map is learned based on the historical trial-and-error experience rather than the visual image.  
In order to further improve the navigation efficiency, a non-local target memory aggregation module is designed to leverage a non-local network to model the intrinsic relationship between the target semantic and the target orientation clues during the navigation process so as to mine the most target-correlated object clues for the navigation decision. 
Extensive experimental results on AI2-Thor and RoboTHOR benchmarks verify the excellent obstacle avoidance and navigation efficiency of our proposed method.
The core source code is available at https://github.co\\m/xwaiyy123/object-navigation.
\end{abstract}





\maketitle

\section{Introduction}
Indoor goal-driven navigation task is a vital task in computer vision and has been widely used in various real-world applications, such as ~\cite{li2021ion,chaplot2020object, luo2022stubborn}. 
Given a predetermined target object, the objective of the navigation task is to enable the agent to automatically locate the target object in unseen indoor settings~\cite{zhang2021hi,duvtnet,dang2022unbiased}. 
Benefiting from the development of the reinforcement learning (RL), most navigation models focus on leveraging RL technique to realize the auto-navigation in a trial-and-error manner. 
Taking a sequence of RGB images obtained during the agent's interaction with the environment as input, a policy network is trained to generate a series of actions that direct the agent toward the target object.
However, due to unknown spatial layout and strong noise interference lying in the unseen scenes, robust indoor goal-driven navigation is still a challenging task. 

\begin{figure*}[h]
  \centering
  \includegraphics[width=18cm]{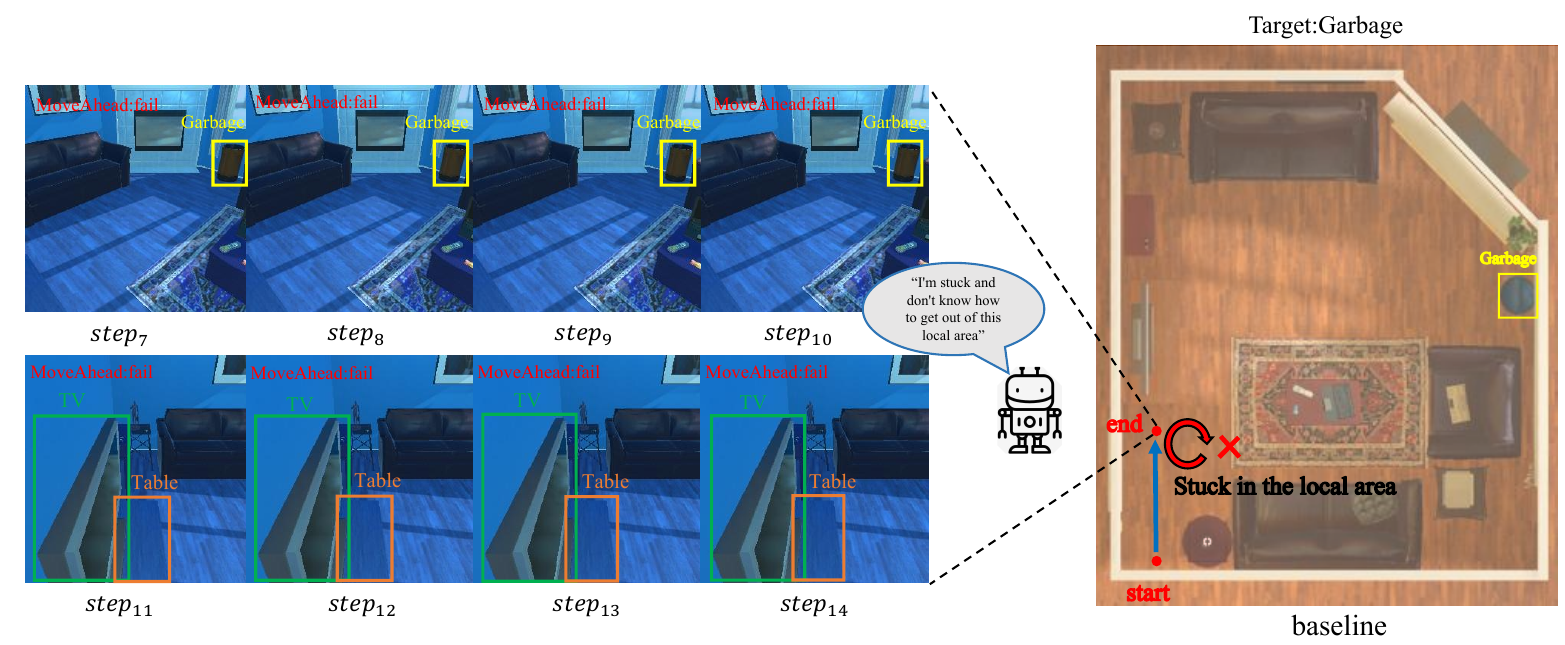}
  \caption{These consecutive frames are selected from an episode, no the obstacle is included in \begin{math} \boldsymbol{step}_7\sim \boldsymbol{step}_{10} \end{math}, the agent believes it can move forward, and does not understand the obstacle distribution in \begin{math} \boldsymbol{step}_{11}\sim \boldsymbol{step}_{14} \end{math}, which still believes that it can move forward.}
  \label{fig1}
\end{figure*}

The popular object navigation models can be categorized into two primary approaches: end-to-end navigation \cite{fukushima2022object, du2020learning, ye2021hierarchical} and BEV (bird’s-eye view) semantics-based navigation \cite{chaplotl11, chaplot2020object}. 
End-to-end navigation methods aim to learn reactive motion decisions based on the perceived environmental information, whereas BEV semantics-based navigation methods focus on utilizing the gathered scene information to construct a BEV (bird's-eye view) scene map for route planning. 
Despite their effectiveness, these methods still struggle to achieve robust obstacle perception (\textit{e.g.}, table and wall) during the navigation process. 
Consequently, the agent is prone to get ``stuck" (becoming immobilized) in a local area, ultimately resulting in navigation failure. 
For instance, if the obstacle distribution around the agent is not included in the image or the model does not recognize the obstacle information in the image, the model might erroneously believe that there are no obstacles in the current direction and repeatedly performs incorrect motion actions, leading to immobilization. 
Figure~\ref{fig1} depicts a typical scenario where the agent has identified the target object, namely garbage, but the obstacle, \textit{i.e.}, a table, is not included in the image. 
The agent presumes that it can approach the target object by moving forward, and repeatedly performs ``moveforward" actions (as seen in \begin{math} step_7\sim step_{10} \end{math}), leading to it getting stuck at its current position.

To mitigate it, in this paper, we propose a novel and effective implicit obstacle map-driven indoor object navigation framework for robust obstacle avoidance during navigation process. 
By encoding ``passable" or ``impassable" direction as the obstacle distribution, the navigation model is expected to precisely recognize the surrounding obstacles so as to help agent choose the correct direction and escape from these obstacles. 
Specifically, our framework consists of two modules, \textit{i.e.}, an implicit obstacle map module and a non-local target memory aggregation module. 
In the implicit obstacle map module, it aims to learn an implicit layout representation that describes the potential obstacle distribution in the local area around the agent through trial-and-error. 
In detail, during the navigation process, when an obstacle is touched in the agent's forward direction, the current direction is impassable and would be encoded as a ``impassable'' vector representation. 
Conversely, if no obstacle is touched, the current direction is passable and thus encoded as a ``passable'' vector representation. 
By aggregating these historical ``impassable'' and ``passable'' vector representations with an MLP network, we can learn an implicit representation about the potential obstacle layout around the agent for obstacle perception. 
Compared to previous visual image-based obstacle identification, our trial-and-error obstacle perception presents significant robustness to obstacle missing in the image or the issue of detection failure. 
In order to further accelerate the approaching speed of the agent towards the target object, the non-local target memory aggregation module employs a non-local module to conduct cross-attention between the target semantic and the collected target orientation clues during the navigation process. 
As such, the vital object clues that are highly correlated with the target object would be mined for navigation decisions, thereby significantly improving the navigation efficiency (that is reducing the path length). 

To summarize, our main contributions are as follows: 

{\bfseries 1)} We propose a novel implicit obstacle map-enhanced navigation framework for robust obstacle avoidance, where the implicit obstacle map is generated through a trial-and-error mechanism, presenting a significant advantage in robustness over the conventional visual image-based obstacle recognition methods. 

{\bfseries 2)} To further improve the navigation efficiency, a non-local target memory aggregation module is proposed to mine the vital target object clues for navigation decisions through the cross-attention mechanism between the target orientation clues and the target semantic. 

{\bfseries 3)} Extensive experimental results on AI2-Thor and RoboTHOR benchmarks verify our significant navigation advantages, in terms of the success ratio and efficiency, over previous methods. 

\section{Related works}
The indoor goal-driven navigation model mainly consists of two parts, namely the feature extractor and policy network, the continuous improvement of these two parts has greatly improved the performance of the navigation model.

The image feature contains a lot of spatial layout information, so the end-to-end navigation model can be built by using the image feature \cite{zhu2017target, mayo2021visual, gadre2022continuous}. 
A large number of experiments have verified that object feature obtained by object detector plays a key role in improving the performance of the navigation model \cite{du2020learning}, 
and more methods tend to use object features to construct navigation model \cite{fukushima2022object, ramakrishnan2022poni, pal2021learning}. 
Du et al. \cite{du2020learning} and Zhang et al. \cite{zhang2021hi} used the relationship between objects to build graphs, graph feature helps the agent quickly find the target object location, which greatly improves the efficiency of the model. 
There are many similar methods \cite{dang2022unbiased, yang2018visual,hu2021agent, li2021ion, ye2021hierarchical, campari2020exploiting}. 
However, Frequent changes of objects in the agent's view can destroy the graph structure and affect the stability of the model.
Some works \cite{savinovsemi, zhuepisodic} built memory features to retain some important image features or object features. 
The transformer \cite{vaswani2017attention, fukushima2022object, fang2019scene} and the graph \cite{savinovsemi, wu2019bayesian} are used to process the memory feature. 
But using the transformer or graph to process a large number of memory features may lead to a large amount of calculation and the inefficiency of the model. 
In our method, the size of the memory is limited, and the non-local aggregation with a simple structure is used for the target memory to reduce the amount of calculation and maintain the performance of the model.

The navigation method based on the semantic map \cite{chaplotl11} is an idea that combines deep learning with traditional SLAM \cite{snavely2008modeling,izadi2011kinectfusion}. 
The construction process of semantic map is divided into two steps \cite{chaplot2020object}, the semantic segmentation model \cite{hu2019acnet,he2017mask,jiang2018rednet} segments objects in the image, then both depth image and object semantic are projected to Bird's Eye View(BEV) to get semantic map. 
The policy network outputs action distribution to complete obstacle avoidance and navigation only based on the semantic map, this method \cite{chaplotl11,chaplot2020object, liang2021sscnav, ramakrishnan2022poni} has a strong dependency on semantic map. 
However, a large amount of projection error and inaccurate representation of spatial layout results in an imprecise semantic map, which can lead to a decrease in the model performance.
The implicit obstacle map we designed can get the obstacle distribution around the agent, its structure is simple, and the computational complexity is small.

Most navigation models combine RNN and Multilayer Perceptron(MLP) to model policy network \cite{zhu2017target,worts,mayo2021visual,zhang2022generative,wu2019bayesian}. 
There are also a few works that directly use CNN \cite{liang2021sscnav,ye2021hierarchical} or MLP \cite{yang2018visual} as policy networks. 
Most of the policy networks are trained by using reinforcement learning, including asynchronous advanced actor critical \cite{zhang2021hi,zhu2017target,fukushima2022object,mnih2016asynchronous} (A3C), proximal policy optimization \cite{maksymets2021thda,schulman2017proximal}(PPO), and Deep Q-learning \cite{liang2021sscnav,ye2021hierarchical,mnih2015human}. 
Very few works \cite{ramakrishnan2022poni} use supervised learning to train models.  
Some models \cite{duvtnet,dang2022unbiased,du2020learning} are trained using both supervised learning and reinforcement learning. 
The models based on semantic map \cite{chaplot2020object,ramakrishnan2022poni,luo2022stubborn} use the traditional Fast Marching Method \cite{sethian1996fast} for path planning and don't need to train.

\section{Method}
\subsection{Problem Setting}
In the context of the indoor object navigation task, given an unseen environment and a predetermined target object, the goal-driven indoor navigation method aims to control the moving of the agent so as to reach the spatial position of the target object. 
We formulate the indoor navigation task as a Markov decision process, consisting of 3 elements: a state set, an action set, and a reward function. 
The state set: observed RGB image; the action set: \begin{math} \label{sects}\mathscr{A} = \{MoveAhead;\, RotateLeft; \,RotateRight;\,LookDown; \,LookUp;\\ \,Done\}  \end{math}. 
During the navigation process, at each time step, given the current state $\boldsymbol{o}_{\boldsymbol{t}} $ and the target object \begin{math} \boldsymbol{g} \end{math}, the agent makes the action decision $\boldsymbol{a}_{\boldsymbol{t}}$ based on the learned policy network $\pi \left(\boldsymbol{a}_{\boldsymbol{t}}|\boldsymbol{o}_{\boldsymbol{t}},\boldsymbol{g} \right)$ and achieves the corresponding reward $\boldsymbol{r}_{\boldsymbol{t}}$. 
The RL algorithm aims to learn the optimal navigation decision so as to maximize the rewards for reaching the location of the target object efficiently. 

\begin{figure*}[h]
  \centering
  \includegraphics[width=12cm]{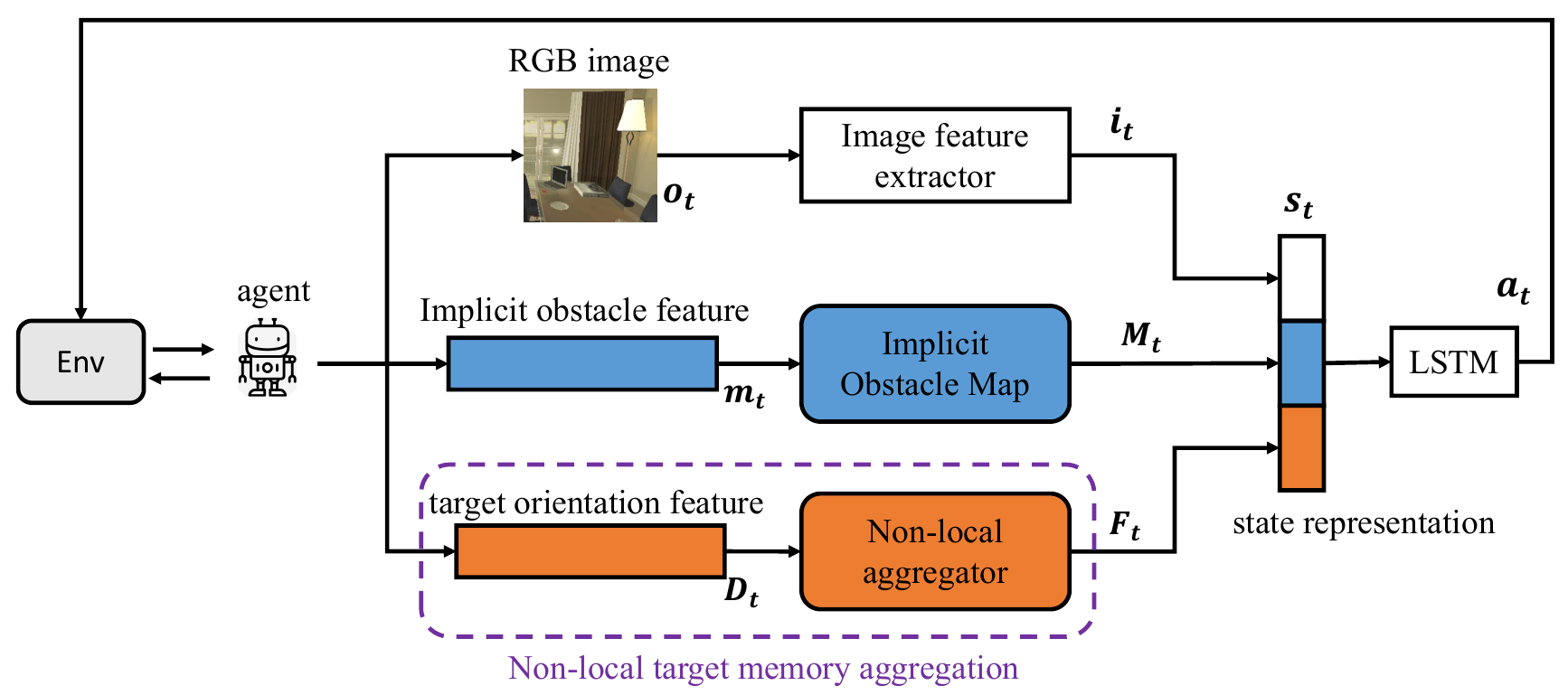}
  \caption{Model overview. The agent obtains RGB image \begin{math} \boldsymbol{o}_{\boldsymbol{t}} \end{math}, implicit obstacle vector feature \begin{math} \boldsymbol{m}_{\boldsymbol{t}} \end{math}, and target orientation feature \begin{math} \boldsymbol{D}_{\boldsymbol{t}} \end{math} when it interacts with the environment. 
  	Image feature extractor outputs image embedding and object feature \begin{math} \boldsymbol{i}_{\boldsymbol{t}} \end{math} (boundingbox, confidence, etc.) based on the  RGB image \begin{math} \boldsymbol{o}_{\boldsymbol{t}} \end{math}. 
  	The implicit obstacle distribution feature \begin{math} \boldsymbol{m}_{\boldsymbol{t}} \end{math} is put into the implicit obstacle map to get the implicit obstacle embedding feature \begin{math} \boldsymbol{M}_{\boldsymbol{t}} \end{math}. 
  	The non-local aggregator outputs target orientation embedding feature \begin{math} \boldsymbol{F}_{\boldsymbol{t}} \end{math} based on the collected target orientation feature \begin{math} \boldsymbol{D}_{\boldsymbol{t}} \end{math}. \begin{math} \boldsymbol{i}_{\boldsymbol{t}} \end{math}, \begin{math} \boldsymbol{M}_{\boldsymbol{t}} \end{math}, and \begin{math} \boldsymbol{F}_{\boldsymbol{t}} \end{math} are concatenated as the state representation \begin{math} \boldsymbol{s}_{\boldsymbol{t}} \end{math}. 
  	Finally, the policy network LSTM outputs the action distribution according to \begin{math} \boldsymbol{s}_{\boldsymbol{t}} \end{math}.}
  \label{fig2}
\end{figure*}

In order to ensure that the agent quickly learns to avoid obstacles and navigate to the target efficiently, we propose two new modules based on the existing feature extraction modules: implicit obstacle map (IOM) and non-local target memory aggregation module (NTMA). 
The overall structure of the designed model is shown in Figure~\ref{fig2}. When the agent interacts with the environment, the image \begin{math} \boldsymbol{o}_{\boldsymbol{t}} \end{math}, the implicit obstacle distribution feature \begin{math} \boldsymbol{m}_{\boldsymbol{t}} \end{math}, and the target orientation feature \begin{math} \boldsymbol{D}_{\boldsymbol{t}} \end{math} are obtained. 
The \begin{math} \boldsymbol{m}_{\boldsymbol{t}} \end{math} and the \begin{math} \boldsymbol{D}_{\boldsymbol{t}} \end{math} are used to build implicit obstacle map and target memory respectively. 
Both the implicit obstacle map embedding feature \begin{math} \boldsymbol{M}_{\boldsymbol{t}} \end{math} and the target memory aggregation embedding feature \begin{math} \boldsymbol{F}_{\boldsymbol{t}} \end{math} outputed by non-local target memory aggregation are used as state representation \begin{math} \boldsymbol{s}_{\boldsymbol{t}} \end{math}. 
Concatenating the image feature and object feature \begin{math} \boldsymbol{i}_{\boldsymbol{t}} \end{math}, the policy network outputs the action distribution \begin{math} \boldsymbol{a}_{\boldsymbol{t}} \end{math} according to the state \begin{math} \boldsymbol{s}_{\boldsymbol{t}} \end{math}. 

\subsection{Implicit obstacle map generation}
In the indoor scene, the obstacle distribution around the agent can be implicitly obtained according to the change of the agent's coordinate when the agent moves forward.
Assuming that the navigation model sends a ``forward" command at the current moment, if the agent's coordinate is not changed, there is an obstacle in the agent's current direction, otherwise, the current direction is passable.

\begin{figure}[h]
  \centering
  \includegraphics[width=\linewidth]{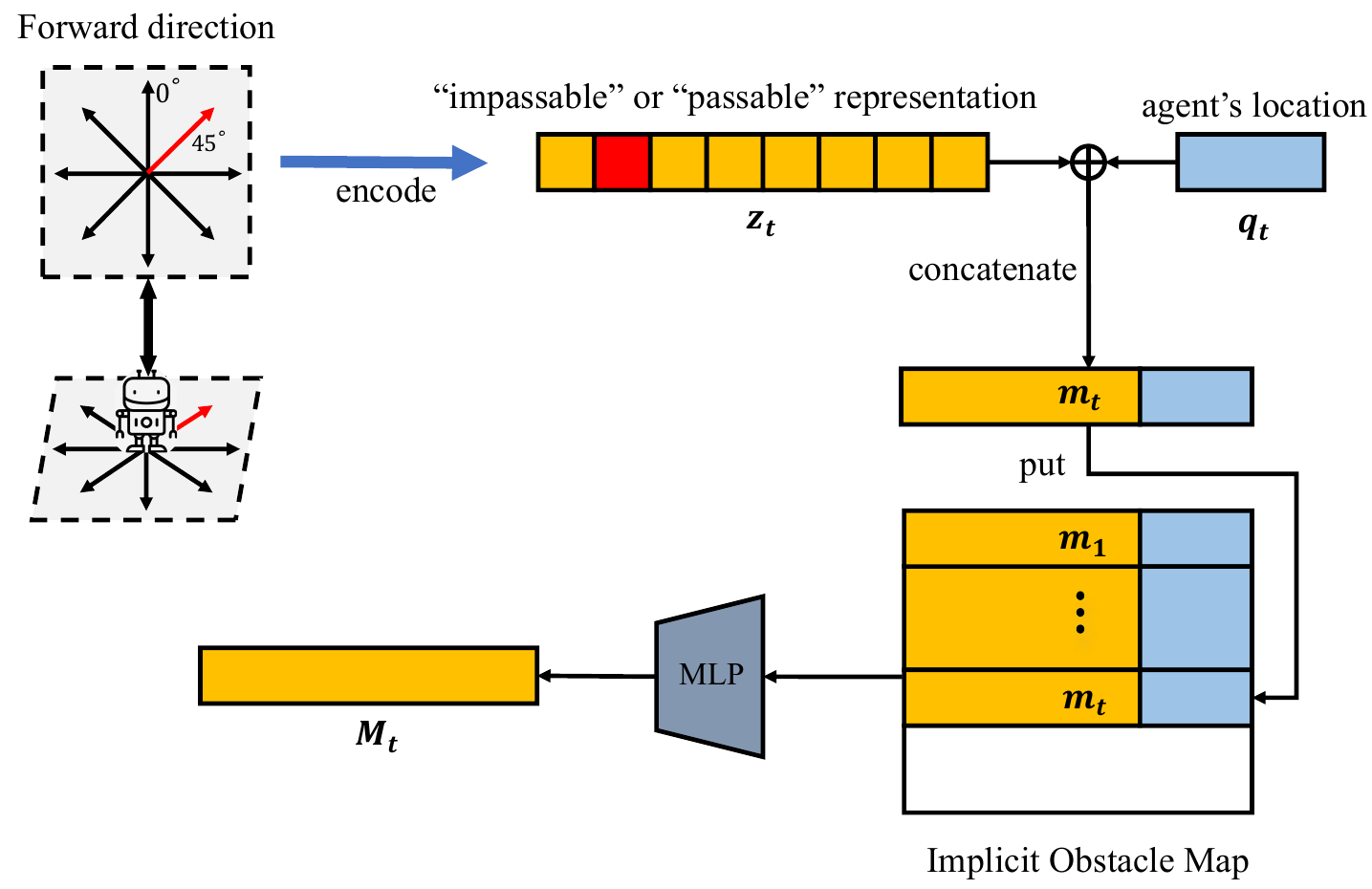}
  \caption{The “impassable” or “passable” representation are encoded as the feature vector \begin{math} \boldsymbol{z}_{\boldsymbol{t}} \end{math}. 
  	\begin{math} \boldsymbol{z}_{\boldsymbol{t}} \end{math} and the agent's coordinates \begin{math} \boldsymbol{q}_{\boldsymbol{t}} \end{math} are concatenated to get the implicit obstacle distribution feature \begin{math} \boldsymbol{m}_{\boldsymbol{t}} \end{math} of the current position. 
  	\begin{math} \boldsymbol{m}_{\boldsymbol{t}} \end{math} is input into the implicit obstacle map. 
  	After the implicit obstacle map is extracted by the MLP, the obstacle distribution embedding feature is obtained.}
  \label{fig3}
\end{figure}

Based on the above characteristic, the implicit obstacle map module is proposed to construct the obstacle distribution around the agent. 
The working process of implicit obstacle map is shown in Figure~\ref{fig3}. For each reachable position, the agent has \begin{math} 8 \end{math} forward directions, and every two adjacent directions differ by \begin{math} 45^{\circ} \end{math}. 
\begin{math} \boldsymbol{z}_{\boldsymbol{t}}\in \mathbb{R} ^{1\times 8} \end{math} indicates whether the 8 directions of the current position are passable. 
The agent is orienting the direction of the red arrow and moves forward.
If the agent's coordinate has not changed, the current direction is impassable, and the corresponding location (red square in Figure~\ref{fig3}) of the vector feature \begin{math} \boldsymbol{z}_{\boldsymbol{t}} \end{math} is set to \begin{math} -1 \end{math}, otherwise, it is set to \begin{math} 1 \end{math}. 
The other locations of \begin{math} \boldsymbol{z}_{\boldsymbol{t}} \end{math} are set to \begin{math} 0 \end{math}. Then the agent's current coordinate \begin{math} \boldsymbol{q}_{\boldsymbol{t}}\in \mathbb{R} ^{1\times 2} \end{math} and \begin{math} \boldsymbol{z}_{\boldsymbol{t}} \end{math} are concatenated to construct the implicit obstacle distribution feature \begin{math} \boldsymbol{m}_{\boldsymbol{t}}\in \mathbb{R} ^{1\times 10} \end{math}. 
\begin{math} \boldsymbol{m}_{\boldsymbol{t}} \end{math} can implicitly represent the obstacle distribution around the agent's current position, and it is also the basic unit of the implicit obstacle map. 
If the current direction is passable, \begin{math} \boldsymbol{m}_{\boldsymbol{t}} \end{math} is put into a fixed-size implicit obstacle map (\begin{math} \boldsymbol{IOM} \in \mathbb{R} ^{e\times 10} \end{math})  (The rest is filled with zeros). 
If the current direction is impassable or the agent's current location has been reached before (The coordinate is same as the previous position), \begin{math} \boldsymbol{z}_{\boldsymbol{i}} \end{math} that included in the previous corresponding feature \begin{math} \boldsymbol{m}_{\boldsymbol{i}} (i\leqslant t-1) \end{math} continues to be encoded.
When the number of the previous implicit features \begin{math} \left\{ \boldsymbol{m}_{\boldsymbol{i}}|i\leqslant t-1 \right\} \end{math} exceeds \begin{math} e \end{math},
\begin{math} e \end{math} previous implicit obstacle distribution features \begin{math} \left\{ \boldsymbol{m}_{\boldsymbol{i}}|i\leqslant t-1 \right\} \end{math} that are closest to the current implicit obstacle distribution feature \begin{math} \boldsymbol{m}_{\boldsymbol{t}} \end{math} are retained.
The features in the implicit obstacle map is then extracted by two linear layers:
\begin{equation}
  \boldsymbol{y_1}=\delta _1\left( Ln_1\left(\boldsymbol{IOM} \right) \right), 
  \label{equ4}
\end{equation}
\begin{equation}
  \boldsymbol{y_2}=\delta _2\left( Ln_2\left( \boldsymbol{y}_{\boldsymbol{1}}^{T} \right) \right) ^T,
  \label{5}
\end{equation}
where \begin{math} \delta _1 \end{math} and \begin{math} \delta _2 \end{math} are activation function \begin{math} Relu \end{math}. 
\begin{math} \boldsymbol{y}_{\boldsymbol{1}}\in \mathbb{R} ^{e\times 32} \end{math} that has a higher dimension is obtained by the linear layer \begin{math} Ln_1 \end{math}, which helps the agent better understand the obstacle distribution. 
The role of \begin{math} Ln_2 \end{math} is to compress the size of the implicit obstacle map embedding representation \begin{math} \boldsymbol{y_1} \end{math} so that the policy network can make full use of the feature information in the \begin{math} \boldsymbol{y_1} \end{math}. 
Ultimately, implicit obstacle map embedding representation \begin{math} \boldsymbol{M_t}= \boldsymbol{y_2}\in \mathbb{R} ^{1\times 32}\end{math} is the embedding representation needed by the policy network. 
Because \begin{math} \boldsymbol{M_t} \end{math} includes the obstacle distribution information around the current agent, the policy network can output reasonable actions to avoid obstacles according to \begin{math} \boldsymbol{M_t} \end{math}.

\subsection{Non-local target memory aggregation}
The agent completes obstacle avoidance by rotating and moving forward. 
The insensitivity for target orientation makes the agent choose an unreasonable rotation direction only based on implicit obstacle map, 
which leads the agent to move away from the target object and creates inefficient navigation.
Therefore, inspired by \cite{dang2022search}, 
we design a non-local target memory aggregation module.
The core idea of the non-local target memory aggregation is to perform a cross-attention between the multiple target orientation features in the target memory and the current target semantic, then target memory is aggregated into a weighted target orientation feature. 
The weighted target orientation feature provides important target orientation information for the agent.

\begin{figure}[h]
  \centering
  \includegraphics[width=\linewidth]{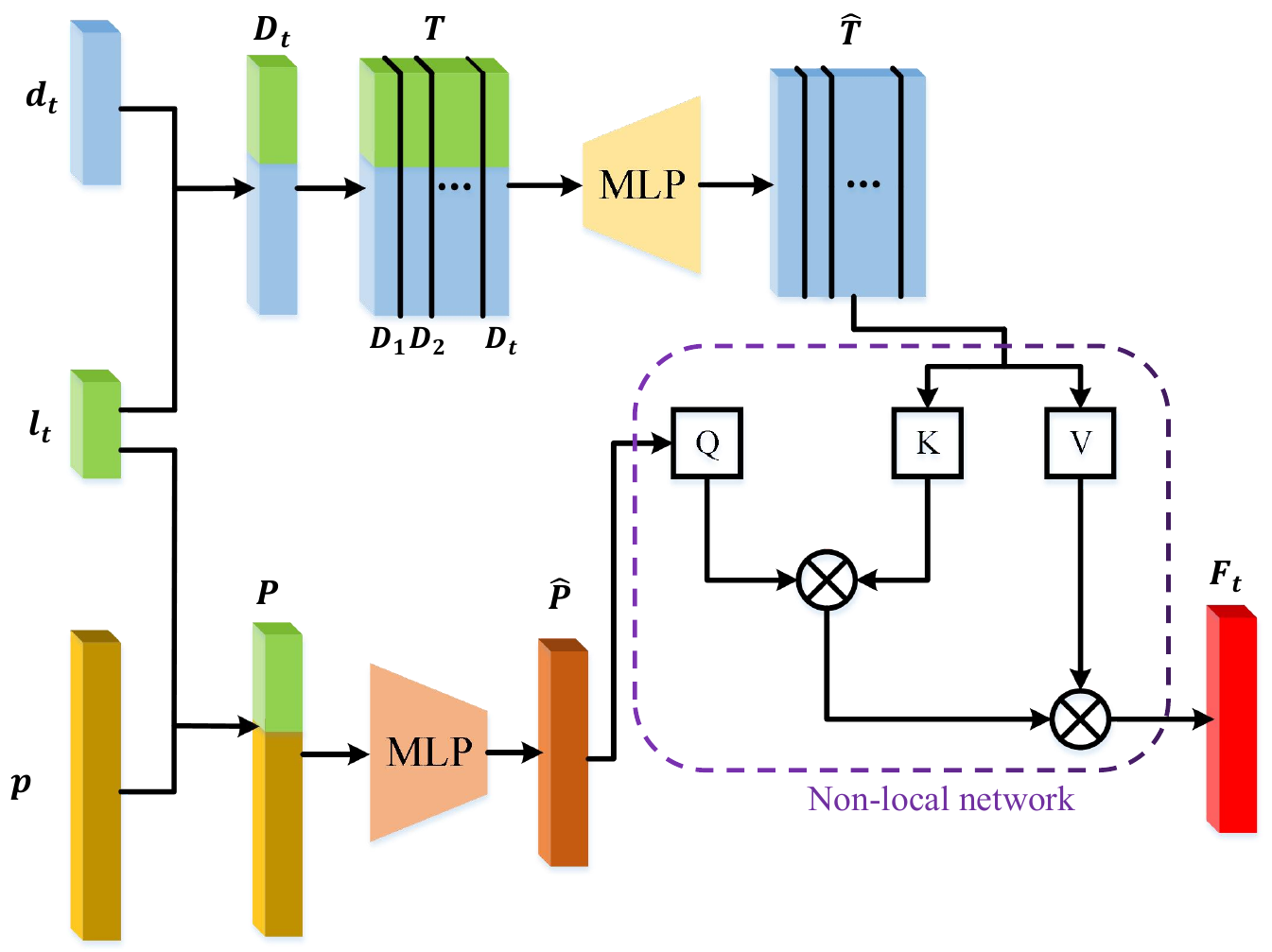}
  \caption{Target orientation features are accumulated to obtain target memory. 
  	The target memory feature and target semantic feature perform non-local aggregation to obtain the target orientation embedding feature.}
  \label{fig4}
\end{figure}

As shown in Figure~\ref{fig4}, for the current state, the target orientation feature \begin{math} \boldsymbol{D_t} \in \mathbb{R} ^{1\times 9} \end{math} is obtained by concatenating object feature \begin{math} \boldsymbol{d_t} \in \mathbb{R} ^{1\times 5} \end{math} (the bounding box and confidence of the target object) obtained by DETR and the agent's pose \begin{math} \boldsymbol{l_t} \in \mathbb{R} ^{1\times 4} \end{math} (plane coordinate \begin{math} \boldsymbol{q_t} =(x,y) \end{math}, yaw angle \begin{math} \boldsymbol{\theta _r} \end{math}, pitch angle \begin{math} \boldsymbol{\theta _h} \end{math}). 
The \begin{math} \boldsymbol{D_t}  \end{math} is then concatenated with previous target orientation features \begin{math} \boldsymbol{D_1} \sim \boldsymbol{D_{t-1}} \end{math} to construct the target memory \begin{math} \boldsymbol{T} \in \mathbb{R} ^{n \times 9} \end{math}, \begin{math} n \leqslant \tau \end{math} is the number of features currently accumulated, and \begin{math} \tau \end{math} is the maximum length of \begin{math} \boldsymbol{T} \end{math}. 
If \begin{math} n> \tau \end{math}, the previous feature \begin{math} \boldsymbol{D_i}  \end{math} that is farthest from \begin{math} \boldsymbol{D_t}  \end{math} will be discarded. 
After expanding the dimension by the two layers of MLP, \begin{math} \boldsymbol{T} \end{math} is converted to \begin{math} \boldsymbol{\widehat{T}} \in \mathbb{R} ^{n \times 32} \end{math}.
The original target semantic \begin{math} \boldsymbol{g} \in \mathbb{R} ^{1 \times k} \end{math} is a one-hot vector (\begin{math} k \end{math} represents the number of object categories in all scenes). 
To keep the target semantics consistent with the current scene features, \begin{math} \boldsymbol{g} \end{math} is transformed into the current target semantic \begin{math} \boldsymbol{p} \in \mathbb{R} ^{1 \times 64} \end{math} by another two layers of MLP. 
The target semantic \begin{math} \boldsymbol{p} \end{math} is then concatenated with the agent's pose \begin{math} \boldsymbol{l_t} \end{math} to enhance target semantic \begin{math} \boldsymbol{p} \end{math} and obtain target semantic feature \begin{math} \boldsymbol{P} \in \mathbb{R} ^{1 \times 68} \end{math}. 
After reducing the dimension by the two layers of MLP, \begin{math} \boldsymbol{P} \end{math} is converted to \begin{math} \boldsymbol{\widehat{P}} \in \mathbb{R} ^{1 \times 32} \end{math}.

In order to get the important target orientation embedding \begin{math} \boldsymbol{F_t} \end{math}, \begin{math} \boldsymbol{P} \end{math} is treated as query, and \begin{math} \boldsymbol{\widehat{T}} \end{math} as key and value, which are then input into the non-local network for cross-attention:
\label{sec4.4}
\begin{equation}
\begin{aligned}
  &\boldsymbol{\widehat{Q}_i}=\boldsymbol{\widehat{P}\,\widehat{W}_{i}^{Q}},\,\,\, \boldsymbol{\widehat{K}_i}=\boldsymbol{\widehat{T}\,\widehat{W}_{i}^{K}},\,\,\,\, \boldsymbol{\widehat{V}}=\boldsymbol{\widehat{T}\,\widehat{W}^V},\\
  &\boldsymbol{\widehat{h}_i}=\mathrm{softmax} \left( {{\boldsymbol{\widehat{Q}_i\widehat{K}_{i}^{T}}}\Bigg/{\sqrt{d_k}}} \right), \,\, \,\,i=1,2,\cdots,\widehat{Nh}, \\
  & \boldsymbol{F_t}=cat\left( \boldsymbol{\widehat{h}_1}, \boldsymbol{\widehat{h}_2},\cdots , \boldsymbol{\widehat{h}_{\widehat{Nh}}} \right) \boldsymbol{\widehat{W}^O \,\widehat{V}},
\end{aligned} 
  \label{equ9}
\end{equation}
where \begin{math} \boldsymbol{\widehat{h}_i} \end{math} is the weight that represent which target orientation features are more important for the current target semantic \begin{math} \boldsymbol{\widehat{P}} \end{math}. 
In \begin{math} \boldsymbol{\widehat{V}} \end{math}, the features that are more correlated with \begin{math} \boldsymbol{\widehat{P}} \end{math} are assigned larger weights, so that \begin{math} \boldsymbol{\widehat{h}_i} \end{math} is used to weight and sum the features in \begin{math} \boldsymbol{\widehat{V}} \end{math}, and target orientation embedding \begin{math} \boldsymbol{F_t} \in \mathbb{R} ^{1\times 32} \end{math} is got by formula~\ref{equ9}. 
Since important target orientation information is included in \begin{math} \boldsymbol{F_t} \end{math}, the agent can choose a more reasonable rotation direction based on \begin{math} \boldsymbol{F_t} \end{math} to approach the target faster and improve navigation efficiency.

\subsection{Navigation Decision Based on RL}
\subsubsection{Policy network}
In the image feature extraction, the image feature \begin{math} \boldsymbol{\widehat{i}_t} \end{math} output by ResNet\cite{he2016deep} and the weighted object feature \begin{math} \boldsymbol{\widehat{r}_t} \end{math}\cite{dang2022unbiased} output by DETR \cite{carion2020end} are obtained, and they are adaptively fused with the implicit obstacle map embedding representation \begin{math} \boldsymbol{M_t}\end{math} and target orientation embedding representation \begin{math} \boldsymbol{F_t} \end{math}. 
The state representation \begin{math} \boldsymbol{s_t} \end{math} is then obtained:
\begin{equation}
  \boldsymbol{s_t}=cat\left( \boldsymbol{\,\widehat{i}_t},\boldsymbol{\widehat{r}_t}, \boldsymbol{M_t}, \boldsymbol{F_t} \right) \boldsymbol{W_a}
  \label{equ10}\,,
\end{equation}
the LSTM\cite{hochreiter1997long} module is treated as a policy network \begin{math}  \pi \left(\boldsymbol{a_t}|\boldsymbol{s_t} \right) \end{math}, and the asynchronous advantage actor-critic(A3C) algorithm\cite{mnih2016asynchronous} is used to train the navigation model.

\subsubsection{Reward Function}
We also design a new reward mechanism to extend the positive impact of implicit obstacle map and non-local target memory aggregation on the model, the detail is as follows:

1) -0.01   \, \,\, \, \,\, Penalty reward for each time step;

2)  0.01   \, \,\, \, \,\,     The agent does not find the object, and the agent moves forward;

3)  0.01    \, \,\, \, \,\,   \, The agent finds the object, and the distance between the agent and the target decreases;
   
4) -0.01   \, \,\, \, \,\,    The agent collides with an obstacle;

5)  0.02  \, \,\, \, \,\, \,     The agent avoids the obstacle;

6)  5  \, \,\, \, \,\, \,\, \,\, \,    Success.

\section{Experiment}
\subsection{Experimental Setting}
\subsubsection{Dataset.}
The AI2-Thor\cite{kolve2017ai2} and the RoboTHOR\cite{deitke2020robothor} are selected to evaluate the performance of our method. 
Because their environment layout is very close to the real indoor scene, many works\cite{yang2018visual, worts, du2020learning, zhang2021hi} use them as the test environment. 
AI2-Thor includes 4 types of room layouts: kitchen, living room, bedroom, and bathroom, each room layout consists of 30 floorplans, of which 20 rooms are used for training, 5 rooms for validation, and 5 rooms for testing. 
RoboTHOR consists of 75 scenes, 60 of which are used for training and 15 for validation. 
The spatial layout of RoboTHOR is more complicated than that of AI2-Thor.

\subsubsection{Evaluation Metrics.} Success rate (SR), success weighted by path length (SPL)\cite{anderson2018evaluation}, and success weighted by action efficiency (SAE) \cite{zhang2021hi} metrics are used to evaluate our method. SR is the success rate of the agent in completing the goal-driven navigation task, and its formula is \begin{math} SR=\frac{1}{K}\sum\nolimits_{i=1}^K{Suc_i} \end{math}, where \begin{math} K \end{math} is the number of episodes, and \begin{math} Suc_i \end{math} is indicates whether the \begin{math} i \end{math}-th episode is successful. 
SPL indicates the efficiency of the agent to complete the task in the successful episodes, its formula is \begin{math} SPL=\frac{1}{K}\sum\nolimits_{i=1}^K{Suc_i}\frac{L_{i}^{*}}{\max \left( L_i,L_{i}^{*} \right)} \end{math}, where \begin{math} L_i \end{math} is the length of the path actually traveled by the agent. 
\begin{math} L_{i}^{*} \end{math} is the optimal path length provided by the simulator. 
SAE represents the proportion of forward actions in successful episodes, which reflects the efficiency of the agent from another perspective, its formula is \begin{math} SAE=\frac{1}{K}\sum\nolimits_{i=1}^K{Suc_i}\frac{\sum\nolimits_{t=0}^T{\mathbb{I} \left( a_{t}^{i}\in \mathscr{A} _{change} \right)}}{\sum\nolimits_{t=0}^T{\mathbb{I} \left( a_{t}^{i}\in \mathscr{A} _{all} \right)}} \end{math}, where \begin{math} \mathbb{I} \end{math} is the indicator function, \begin{math} a_{t}^{i} \end{math} is the  agent’s action at time \begin{math} t \end{math} in episode \begin{math} i \end{math}. \begin{math} \mathscr{A} _{all} \end{math} is action space \begin{math} \mathscr{A} \end{math} in section~\ref{sects}. 
\begin{math} \mathscr{A} _{change} \end{math} represents move ahead.

\subsubsection{Implementation Details.} Our model is trained by 14 workers on 2 RTX 2080Ti Nvidia GPUs. 
Similar to the previous method, we first use action label samples for pre-training and then use reinforcement learning for formal training. 
The total of the agent interacting with the environment is 3M episodes. 
The target confidence threshold \begin{math} C \end{math} that the agent can find the target is set to 0.4.
In the Non-local Target Memory Aggregation, the dropout of attention is 0.1, the number of head is 4. 
The learning rate of the optimizer used to update the model parameters is \begin{math} 10^{-4} \end{math}. 
For evaluation, our test results are all experimented with 3 times and then taken the average.
We show results for all targets (ALL) and a subset of targets (\begin{math} L\geqslant 5 \end{math}) whose optimal trajectory length is longer than 5.

\begin{figure*}[h]
  \centering
  \includegraphics[width=\linewidth]{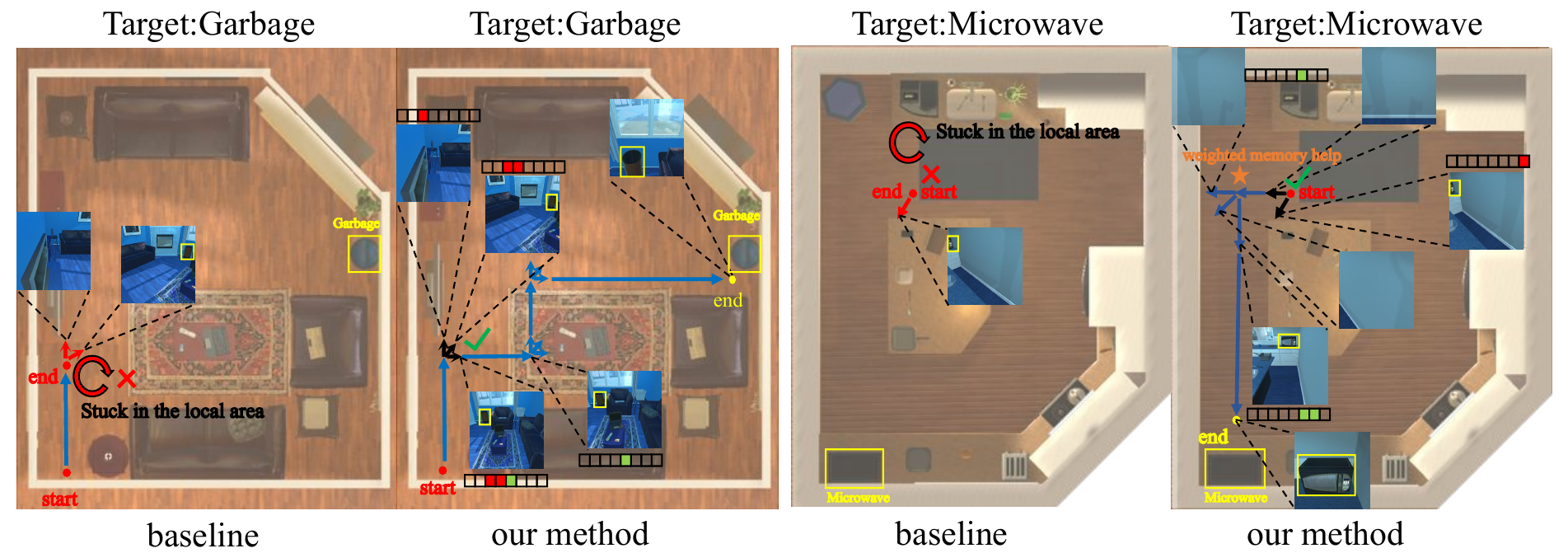}
  \caption{Visualization of the testing process. 
  	The target is selected by the yellow box. 
  	Some key frames are selected to illustrate the working process of our method.}
  \label{fig7}  
\end{figure*}

\subsection{Comparisons to the State-of-the-art}
To demonstrate the superiority of our method, our method is compared with some recent methods. The results of the comparative experiments are shown in Table~\ref{tab:tab1}.

\begin{table*}
  \caption{Comparisons with the state-of-the-art methods on the AI2-Thor/RoboTHOR datasets}
  \label{tab:tab1} 
  \resizebox{10cm}{!}{  
  \begin{tabular}{ c|ccc|ccc|c }
    \toprule[1pt]
   \multirow{2}{*}{Method} & \multicolumn{3}{c|}{ALL (\begin{math} \% \end{math})}& \multicolumn{3}{c}{$ L\geqslant5 $ (\begin{math} \% \end{math})}& \multirow{2}{1.05cm}{Episode Time (s)}\\  
    \cmidrule{2-7} 
    & SR& SPL& SAE & SR& SPL& SAE\\
    \midrule[1pt]
    SP\cite{yang2018visual} & 62.59/23.73 & 37.91/15.94& 24.55/17.81& 52.36/19.47& 33.61/14.27 & 22.67/15.39 & 0.33/1.37\\
    SAVN\cite{worts} & 63.34/25.82 & 37.16/16.68& 20.55/15.69& 52.23/20.55& 34.94/14.37 & 23.15/15.79& 0.31/1.40\\
    SA\cite{mayo2021visual} & 66.51/27.54 & 38.35/17.37& 21.71/16.35& 54.53/21.53& 34.34/15.79 & 25.58/16.09& 0.31/\textbf{1.29}\\
    ORG\cite{du2020learning} & 67.84/29.87 & 36.94/18.34& 26.18/16.78& 58.58/22.43 & 35.61/17.58 & 27.34/15.23& \textbf{0.25}/1.30 \\
    HOZ\cite{zhang2021hi} & 68.49/30.13 & 37.51/18.84& 25.64/17.63& 60.94/23.46& 36.57/18.50 & 27.69/18.33 & 0.30/1.40\\
    VTNet\cite{duvtnet} & 72.37/34.03 & 45.47/20.31 &30.35/19.44& 64.56/30.19& 44.37/19.43 & 31.43/20.63 & 0.34/1.42 \\
    DOA\cite{dang2022unbiased} & 78.84/37.66 &  44.27/21.46& 31.15/20.53& 72.38/32.17& 45.51/20.73 & 36.02/21.29& 0.37/1.37\\
    OMT\cite{fukushima2022object} & 70.54/33.81 &  30.43/19.31& 26.14/18.16& 59.78/29.47& 26.31/19.63 & 24.37/19.14& 0.68/1.53\\
    SSCNav\cite{liang2021sscnav} & 76.28/37.06 & 24.43/13.88& 26.58/16.34& 68.11/31.24& 29.93/14.35 & 24.97/15.34 & 1.46/4.96\\ 
    PONI\cite{ramakrishnan2022poni} & 79.64/38.73 & 35.60/14.26& 28.19/17.35& 73.28/33.46& 36.37/15.34 & 29.37/18.76& 1.83/5.04\\ 
	DAT\cite{dang2022search} & 80.95/39.82 & 46.02/25.34& 31.69/22.37& 73.86/34.33& 46.48/20.47 & 35.91/23.15& 0.32/1.34\\
   \textbf{Our} & \textbf{82.99}/\textbf{42.07}  & \textbf{47.40}/\textbf{27.47}&  \textbf{33.04}/\textbf{24.42}&  \textbf{77.95}/\textbf{37.26}& \textbf{48.78}/\textbf{22.23}  &  \textbf{37.69}/\textbf{25.04}&  0.38/1.35\\ 
  \bottomrule[1pt]
    \end{tabular}
}
\end{table*}
It can be seen that the performance of our method shows a large advantage. 
Importantly, our method outperforms current SOTA method (DAT) with the gains of  2.04/2.25, 1.38/2.13, 1.35/2.05 (AI2-Thor/RoboTHOR, \begin{math} \% \end{math}) in SR, SPL and SAE, this result demonstrates the effectiveness and efficiency of our proposed method. 
Compared with other end-to-end methods, although the semantic map-based methods show greater advantages in SR, these methods spend a lot of time exploring environment and constructing semantic map, which leads to their poor performance on SPL and SAE. 
And semantic map-based methods take much longer episode time (s), which is unacceptable in practical tasks. 
Although our method does not show a great advantage in terms of episode time (s), the episode time that the model takes is still acceptable. 
Moreover, our method obtains the gains of 3.35/3.34 (AI2-Thor/RoboTHOR, \begin{math} \% \end{math}) in SR over the semantic map-based SOTA method PONI, which shows that the semantic map built on the depth image does not show great advantage due to noise interference, and also verifies that our method has good robustness.
 
\subsection{Ablation Experiments}
\subsubsection{Baseline.}
Our baseline is composed of an image feature extractor and target memory. 
As shown in Figure~\ref{fig2}, the resnet18 and the DETR are included in the image feature extractor, and the object features obtained by DETR are assigned learnable weights \cite{dang2022unbiased}. 
The construction process of the target memory is detailed in Section~\ref{sec4.4}. 
All the target orientation features in the target memory are averaged to obtain the target orientation embedding feature. 
The traditional sparse reward is used to train the model.

In order to study the impact of implicit obstacle map (IOM), non-local target memory aggregation (NTWA), and reward mechanism (RM) on the model, a series of ablation experiments are designed to demonstrate the effectiveness of these modules. 
The results of the ablation experiments on the AI2-Thor are shown in Table~\ref{tab:tab2}.
\begin{table}
  \caption{Ablation experiment results for IOM, NTWA, and RM}
  \label{tab:tab2}
  \resizebox{\linewidth}{!}{  
  \begin{tabular}{ ccc|ccc|ccc }
    \toprule[1pt] 
  \multirow{2}{*}{IOM}& \multirow{2}{*}{NTWA} &\multirow{2}{*}{RM} & \multicolumn{3}{c|}{ALL (\begin{math} \% \end{math})}& \multicolumn{3}{c}{$ L\geqslant5 $ (\begin{math} \% \end{math})}\\  
    \cmidrule{4-9} 
    & & & SR& SPL& SAE & SR& SPL& SAE\\
    \midrule[1pt]
    &  &  & 78.51& 46.50& 30.09 & 71.45 & 46.42 & 34.07\\
    \midrule[1pt]
    \Checkmark &  &  & 81.04& 46.61& 31.58 & 74.25 & 46.88 & 35.43\\
     & \Checkmark &  & 80.24& \textbf{47.49} & 30.48 & 73.11 & 47.78 & 34.87\\
     &  & \Checkmark & 78.72 & 46.14 & 28.93 & 72.05 & 46.46 & 33.96\\
    \Checkmark & \Checkmark &  & 82.23& 46.93 & 31.22& 76.97 & 47.35 & 35.52 \\
    \Checkmark &  & \Checkmark & 81.87& 47.46& 32.54 & 75.83 & 48.28 & 35.39\\
    & \Checkmark & \Checkmark & 80.59 & 47.48& 31.20 & 73.17 & 47.53 & 34.88\\
    \Checkmark & \Checkmark & \Checkmark &  \textbf{82.99}  & 47.40&  \textbf{33.04}&  \textbf{77.95}& \textbf{48.78}  &  \textbf{37.69}\\
  \bottomrule[1pt]
    \end{tabular}
}
\end{table}
\subsubsection{Implicit obstacle map.}
Compared with the baseline, the performances of the IOM-based model are improved by 2.53/2.8, 1.49/\\1.36 (ALL/\begin{math} L\geqslant 5 \end{math}, \begin{math} \% \end{math}) in SR and SAE, which indicates that IOM does help the model improve performance. 
However, the improvement of the model in SPL is very small, which shows that the IOM-based model may avoid obstacles with the help of IOM, but the agent does not choose the optimal path to approach the target object. 
After introducing the RM on the IOM-based model, the SPL of the model is improved by 0.96/1.86 (ALL/\begin{math} L\geqslant 5 \end{math}, \begin{math} \% \end{math}), which indicates that the RM increases the efficiency of IOM-based model. 
Most importantly, after using NTWA to process the target memory, the SR of the IOM-based model has been greatly improved. 
Compared with the baseline, SR is improved by 3.72/5.52 (ALL/\begin{math} L\geqslant 5 \end{math}, \begin{math} \% \end{math}). 
Since NTWA assigns large weights to the important target orientation features, the cooperation between IOM and important target orientation features greatly improves the performance of the model. 
The IOM-NTWA-based model also encounters the same problem: compared with the IOM-based model, the performance improvement of SRL is very small, only  0.43/0.93 (ALL/\begin{math} L\geqslant 5 \end{math}, \begin{math} \% \end{math}). 
After adding RM on the basis of IOM and NTWA, the performance of IOM-NTWA-based model has been greatly improved. 
Compared with the baseline, the IOM-NTWA-RM-based model is improved by 4.48/6.5, 0.9/2.36, 2.95/3.62 (ALL/\begin{math} L\geqslant 5 \end{math}, \begin{math} \% \end{math}) in SR, SPL, and SAE, which proves that the introduction of RM makes the cooperation between IOM and NTWA more perfect, and allows our model to outperform existing SOTA methods.

\subsubsection{Non-local Target Memory Aggregation.}
After introducing the NTWA in the baseline, SR, SPL, and SAE are improved by 1.73/1.66, 0.99/1.36, 0.39/0.8 (ALL/\begin{math} L\geqslant 5 \end{math}, \begin{math} \% \end{math}), which shows that the agent can understand important target orientation information through NTWA, and the ability of navigation is improved. 
And the experimental results of the NTWA-based model, IOM-NTWA-based model, and IOM-NTWA-RM-based model show that the role of NTWA is irreplaceable. 
It can be seen from Table~\ref{tab:tab2} that RM has very little promotion effect on NTWA, the SR and SAE are only improved by 0.35/0.06, 0.72/0.01 (ALL/\begin{math} L\geqslant 5 \end{math}, \begin{math} \% \end{math}), and even SPL shows a negative growth. 
This shows that RM is more helpful to IOM than to NTWA. 

\subsubsection{Reward Mechanism.}
From the results in Table~\ref{tab:tab2}, it can be seen that only using RM does not greatly improve the performance of the model. 
Compared with the baseline, SR only improves by 0.1. 
In the absence of IOM and NTWA, the agent cannot understand the reward signal given by RM. 
Our proposed RM can only further amplify the positive influence of IOM and NTWA in the navigation model and cannot affect the performance of the model alone.

\subsection{Importance Analysis of IOM and NTWA}
\subsubsection{Irreplaceability of IOM and NTWA}
In order to demonstrate that the IOM and NTWA are important and irreplaceable, some other schemes are introduced in the baseline and then compared with the IOM-based model and NTWA-based model. 
Comparison experiments on the AI2-Thor are shown in Table~\ref{tab:tab3} and Table~\ref{tab:tab4}.

\begin{table}
  \caption{Comparison experiment results for IOM, TPN, and GM}
  \label{tab:tab3}
  \begin{tabular}{ c|ccc|ccc }
    \toprule[1pt] 
  \multirow{2}{*}{Method} & \multicolumn{3}{c|}{ALL (\begin{math} \% \end{math})}& \multicolumn{3}{c}{$ L\geqslant5 $ (\begin{math} \% \end{math})}\\  
    \cmidrule{2-7} 
    & SR& SPL& SAE & SR& SPL& SAE\\
    \midrule[1pt] 
    baseline & 78.51& 46.50& 30.09 & 71.45 & 46.42 & 34.07\\
    GM & 76.07& 43.99& 28.34 & 66.62 & 42.61 & 31.62\\
    TPN & 79.63& 45.02& 29.00 & 72.51 & 45.74 & 34.15\\
    IOM & \textbf{81.04} & \textbf{46.61}& \textbf{31.58} & \textbf{74.25} & \textbf{46.88} & \textbf{35.43}\\
  \bottomrule[1pt]
    \end{tabular}
\end{table}

For the IOM, {\bfseries 1)} Tentative policy network (TPN) \cite{du2020learning} is introduced in the baseline for comparative experiments. 
TPN is to solve the problem that the agent is stuck in the deadlock state, which is similar to the problem that IOM needs to solve. 
{\bfseries 2)} The grid map(GM) \cite{luo2022stubborn} is used to represent the free space of the agent. 
GM directly expresses the passable area of the agent to help the agent avoid obstacles. 
In our work, the size of GM is set to 40*40, and the CNN is used to extract the features of GM. 
The comparison results of the TPN-based method, the GM-based method, and the IOM-based method are shown in Table~\ref{tab:tab3}. 
It can be seen that the IOM-based method exceeds the TPN-based method and the GM-based method in all aspects of evaluation metrics. 
By the way, TPN is also a method that focuses on choosing the appropriate action from the experience pool. 
The pseudo-label actions output by TPN may still not be able to help the agent get out of the deadlock state. 
The IOM-based method can continuously correct errors according to the implicit obstacle distribution, and then completes obstacle avoidance, the model remains robust even in unseen environments.
As can be seen from Table~\ref{tab:tab3}, the layout information included in GM is too sparse, making the agent difficult to obtain useful obstacle information, so the performance of the GM-Based model is not as good as that of the IOM-Based model.

For the NTWA, {\bfseries 1)} baseline. In the baseline, all the target orientation features in the target memory are averaged, which is also a method to extract features. 
{\bfseries 2)} Target-Aware Multi-Scale Aggregator (TAMSA) \cite{dang2022search}, TAMSA uses multi-scale one-dimensional convolutions to extract target memory features. The results of the experiment are shown in Table~\ref{tab:tab4}. 
\begin{table}
  \caption{Comparison experiment results for NTWA, baseline, and TAMSA}
  \label{tab:tab4}
  \begin{tabular}{ c|ccc|ccc }
    \toprule[1pt] 
  \multirow{2}{*}{Method} & \multicolumn{3}{c|}{ALL (\begin{math} \% \end{math})}& \multicolumn{3}{c}{$ L\geqslant5 $ (\begin{math} \% \end{math})}\\  
    \cmidrule{2-7} 
    & SR& SPL& SAE & SR& SPL& SAE\\
    \midrule[1pt] 
    baseline & 78.51& 46.50& 30.09 & 71.45 & 46.42 & 34.07\\
    TAMSA & 79.22& 46.48& 30.21 & 72.65 & 46.53 & 34.62\\
    NTWA & \textbf{80.24} & \textbf{47.49}& \textbf{30.48} & \textbf{73.11} & \textbf{47.78} & \textbf{34.87}\\
  \bottomrule[1pt]
    \end{tabular}
\end{table}
 It can be seen that the performance of the NTWA-based method is better than the baseline and TAMSA-based method, which shows that assigning reasonable weights to the important target orientation features improve the ability of the agent locating the target. 
 By the way, for the initial stage of an episode, in order to ensure that the size of the target memory is fixed, the target memory embedding features extracted by TAMSA include too many ``0", which limits the navigation capabilities of the model. 
 Since attention is used in the NTWA-based method, there is no need to fill the target memory with ``0" to fix its size, which ensures the reliability of embedding features.
\begin{figure}[h]
  \centering
  \includegraphics[width=6cm]{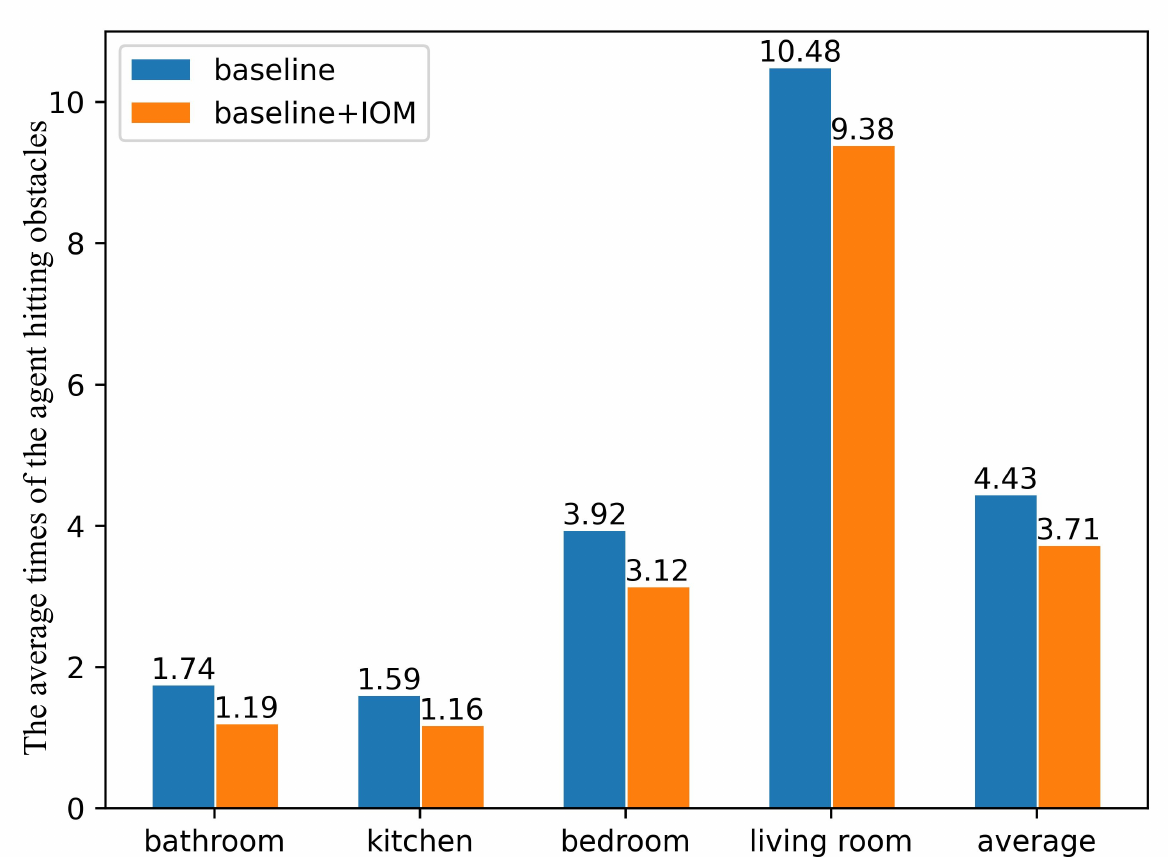}
  \caption{
  The figure represents the average times of the agent based on the different methods colliding with obstacles.}
  \label{fig5}  
\end{figure}

\begin{figure}[h]
  \centering
  \includegraphics[width=8cm]{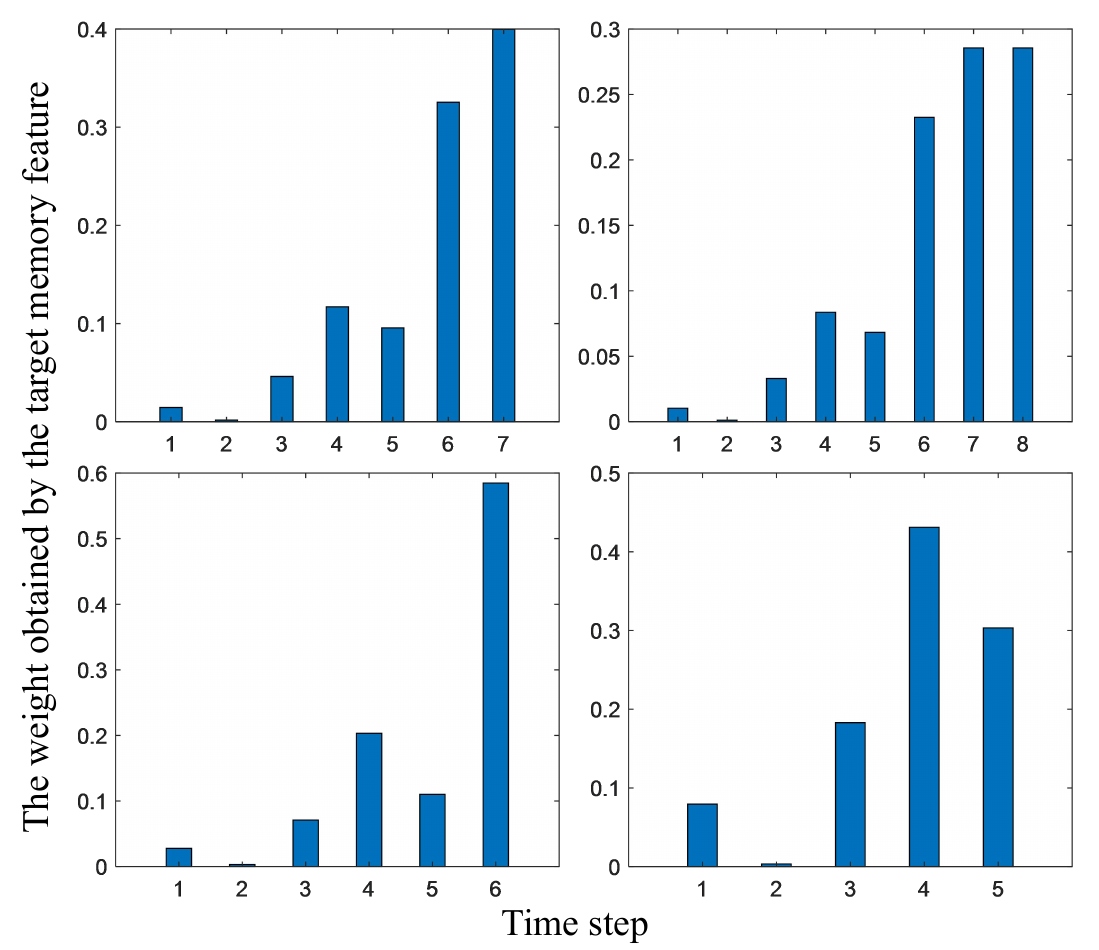}
  \caption{
  	The figure represents the weights assigned by NTWA to different target orientation features.
  }
  \label{fig6}  
\end{figure}

\subsubsection{Does IOM help the agent improve the obstacle avoidance ability?}
If the IOM helps the agent to avoid obstacles, there should be two obvious phenomena: the rotation times of the agent are reduced; and the times of the agent colliding with obstacles in the scene are reduced. 
From Table~\ref{tab:tab3}, the SAE of the IOM-based method is 1.49/1.36 (ALL/\begin{math} L\geqslant 5 \end{math}, \begin{math} \% \end{math}) higher than the baseline, which shows the frequency of agent's rotation is reduced. For the 4 test room layouts(AI2-Thor), the average times of the IOM-base agent and baseline-base agent colliding with obstacles are shown in Figure~\ref{fig5}. 
The times of the agent colliding with obstacles in the test scenes are also reduced. 
In a word, IOM does help the agent improve the ability to avoid obstacles.

\subsubsection{How does the NTWA assign weights to target memory?}
Taking an episode as an example, the weights assigned by NTWA to the target orientation features are shown in Figure~\ref{fig6}. 
The larger the value of abscissa in the figure, the closer the time interval between the target orientation features and the current state. 
As can be seen from the figure, NTWA prefers to assign larger weights to the target orientation features that are closer to the current state, which indicates that the features that are closer to the current state are more important.

\subsection{Qualitative Analysis}
Two typical cases are chosen for visualization in Figure~\ref{fig7}. 
The target of the first case is the ``Garbage". 
Since the baseline model cannot obtain the obstacle distribution information from the image, the agent is stuck in front of the table, failing in the navigation task. 
In our method, the implicit obstacle map encodes the obstacle distribution(Red squares indicate impassable directions, green squares indicate passable directions), so the agent quickly finds the passable direction and avoids the table, then navigates to the target object. 
The target of the second case is ``Microwave". 
Because the agent in the baseline cannot understand the surrounding obstacle distribution and stops at the initial position. 
Our method successfully avoids obstacles with the help of the implicit obstacle map. 
More importantly, since the agent obtains target orientation information by the non-local target memory aggregation, it finds the correct location of the target object when the target disappears in the its view, and completes the navigation task excellently.

\section{Conclusion}
In this paper, we first proposed an implicit obstacle map module to help the agent avoid obstacles, a non-local target memory aggregation was also designed to obtain target orientation embedding features and help the agent locate the target, and a new reward mechanism was then designed to maximize the potential of the implicit obstacle map and non-local target memory aggregation. 
A large number of experimental results showed that our model not only guided the agent well to avoid obstacles, but also improved the agent's navigation efficiency under the cooperation of the implicit obstacle map, non-local target memory aggregation, and new reward mechanism.



\begin{acks}
The authors would like to thank reviewers for their detailed comments and instructive suggestions.
This work was supported by the National Science Fund of China (Grant Nos. 62106106).
\end{acks}


\bibliographystyle{ACM-Reference-Format}
\balance
\bibliography{sample-sigconf}


\begin{thebibliography}{43}


\ifx \showCODEN    \undefined \def \showCODEN     #1{\unskip}     \fi
\ifx \showDOI      \undefined \def \showDOI       #1{#1}\fi
\ifx \showISBNx    \undefined \def \showISBNx     #1{\unskip}     \fi
\ifx \showISBNxiii \undefined \def \showISBNxiii  #1{\unskip}     \fi
\ifx \showISSN     \undefined \def \showISSN      #1{\unskip}     \fi
\ifx \showLCCN     \undefined \def \showLCCN      #1{\unskip}     \fi
\ifx \shownote     \undefined \def \shownote      #1{#1}          \fi
\ifx \showarticletitle \undefined \def \showarticletitle #1{#1}   \fi
\ifx \showURL      \undefined \def \showURL       {\relax}        \fi
\providecommand\bibfield[2]{#2}
\providecommand\bibinfo[2]{#2}
\providecommand\natexlab[1]{#1}
\providecommand\showeprint[2][]{arXiv:#2}

\bibitem[Anderson et~al\mbox{.}(2018)]%
        {anderson2018evaluation}
\bibfield{author}{\bibinfo{person}{Peter Anderson}, \bibinfo{person}{Angel
  Chang}, \bibinfo{person}{Devendra~Singh Chaplot}, \bibinfo{person}{Alexey
  Dosovitskiy}, \bibinfo{person}{Saurabh Gupta}, \bibinfo{person}{Vladlen
  Koltun}, \bibinfo{person}{Jana Kosecka}, \bibinfo{person}{Jitendra Malik},
  \bibinfo{person}{Roozbeh Mottaghi}, \bibinfo{person}{Manolis Savva},
  {et~al\mbox{.}}} \bibinfo{year}{2018}\natexlab{}.
\newblock \showarticletitle{On evaluation of embodied navigation agents}.
\newblock \bibinfo{journal}{\emph{arXiv preprint arXiv:1807.06757}}
  (\bibinfo{year}{2018}).
\newblock


\bibitem[Campari et~al\mbox{.}(2020)]%
        {campari2020exploiting}
\bibfield{author}{\bibinfo{person}{Tommaso Campari}, \bibinfo{person}{Paolo
  Eccher}, \bibinfo{person}{Luciano Serafini}, {and} \bibinfo{person}{Lamberto
  Ballan}.} \bibinfo{year}{2020}\natexlab{}.
\newblock \showarticletitle{Exploiting scene-specific features for object goal
  navigation}. In \bibinfo{booktitle}{\emph{European Conference on Computer
  Vision}}. \bibinfo{pages}{406--421}.
\newblock


\bibitem[Carion et~al\mbox{.}(2020)]%
        {carion2020end}
\bibfield{author}{\bibinfo{person}{Nicolas Carion}, \bibinfo{person}{Francisco
  Massa}, \bibinfo{person}{Gabriel Synnaeve}, \bibinfo{person}{Nicolas
  Usunier}, \bibinfo{person}{Alexander Kirillov}, {and} \bibinfo{person}{Sergey
  Zagoruyko}.} \bibinfo{year}{2020}\natexlab{}.
\newblock \showarticletitle{End-to-end object detection with transformers}. In
  \bibinfo{booktitle}{\emph{European Conference on Computer Vision}}.
  \bibinfo{pages}{213--229}.
\newblock


\bibitem[Chaplot et~al\mbox{.}(2019)]%
        {chaplotl11}
\bibfield{author}{\bibinfo{person}{Devendra~Singh Chaplot},
  \bibinfo{person}{Dhiraj Gandhi}, \bibinfo{person}{Saurabh Gupta},
  \bibinfo{person}{Abhinav Gupta}, {and} \bibinfo{person}{Ruslan
  Salakhutdinov}.} \bibinfo{year}{2019}\natexlab{}.
\newblock \showarticletitle{Learning to explore using active neural SLAM}. In
  \bibinfo{booktitle}{\emph{International Conference on Learning
  Representations}}.
\newblock


\bibitem[Chaplot et~al\mbox{.}(2020)]%
        {chaplot2020object}
\bibfield{author}{\bibinfo{person}{Devendra~Singh Chaplot},
  \bibinfo{person}{Dhiraj~Prakashchand Gandhi}, \bibinfo{person}{Abhinav
  Gupta}, {and} \bibinfo{person}{Russ~R Salakhutdinov}.}
  \bibinfo{year}{2020}\natexlab{}.
\newblock \showarticletitle{Object goal navigation using goal-oriented semantic
  exploration}.
\newblock \bibinfo{journal}{\emph{Advances in neural information processing
  systems}}  \bibinfo{volume}{33} (\bibinfo{year}{2020}),
  \bibinfo{pages}{4247--4258}.
\newblock


\bibitem[Dang et~al\mbox{.}(2022a)]%
        {dang2022unbiased}
\bibfield{author}{\bibinfo{person}{Ronghao Dang}, \bibinfo{person}{Zhuofan
  Shi}, \bibinfo{person}{Liuyi Wang}, \bibinfo{person}{Zongtao He},
  \bibinfo{person}{Chengju Liu}, {and} \bibinfo{person}{Qijun Chen}.}
  \bibinfo{year}{2022}\natexlab{a}.
\newblock \showarticletitle{Unbiased directed object attention graph for object
  navigation}. In \bibinfo{booktitle}{\emph{Proceedings of the 30th ACM
  International Conference on Multimedia}}. \bibinfo{pages}{3617--3627}.
\newblock


\bibitem[Dang et~al\mbox{.}(2022b)]%
        {dang2022search}
\bibfield{author}{\bibinfo{person}{Ronghao Dang}, \bibinfo{person}{Liuyi Wang},
  \bibinfo{person}{Zongtao He}, \bibinfo{person}{Shuai Su},
  \bibinfo{person}{Chengju Liu}, {and} \bibinfo{person}{Qijun Chen}.}
  \bibinfo{year}{2022}\natexlab{b}.
\newblock \showarticletitle{Search for or Navigate to? Dual Adaptive Thinking
  for Object Navigation}.
\newblock \bibinfo{journal}{\emph{arXiv preprint arXiv:2208.00553}}
  (\bibinfo{year}{2022}).
\newblock


\bibitem[Deitke et~al\mbox{.}(2020)]%
        {deitke2020robothor}
\bibfield{author}{\bibinfo{person}{Matt Deitke}, \bibinfo{person}{Winson Han},
  \bibinfo{person}{Alvaro Herrasti}, \bibinfo{person}{Aniruddha Kembhavi},
  \bibinfo{person}{Eric Kolve}, \bibinfo{person}{Roozbeh Mottaghi},
  \bibinfo{person}{Jordi Salvador}, \bibinfo{person}{Dustin Schwenk},
  \bibinfo{person}{Eli VanderBilt}, \bibinfo{person}{Matthew Wallingford},
  {et~al\mbox{.}}} \bibinfo{year}{2020}\natexlab{}.
\newblock \showarticletitle{Robothor: An open simulation-to-real embodied ai
  platform}. In \bibinfo{booktitle}{\emph{Proceedings of the IEEE/CVF
  Conference on Computer Vision and Pattern Recognition}}.
  \bibinfo{pages}{3164--3174}.
\newblock


\bibitem[Du et~al\mbox{.}(2020a)]%
        {du2020learning}
\bibfield{author}{\bibinfo{person}{H Du}, \bibinfo{person}{X Yu}, {and}
  \bibinfo{person}{L Zheng}.} \bibinfo{year}{2020}\natexlab{a}.
\newblock \showarticletitle{Learning Object Relation Graph and Tentative Policy
  for Visual Navigation}. In \bibinfo{booktitle}{\emph{European Conference on
  Computer Vision}}.
\newblock


\bibitem[Du et~al\mbox{.}(2020b)]%
        {duvtnet}
\bibfield{author}{\bibinfo{person}{Heming Du}, \bibinfo{person}{Xin Yu}, {and}
  \bibinfo{person}{Liang Zheng}.} \bibinfo{year}{2020}\natexlab{b}.
\newblock \showarticletitle{VTNet: Visual transformer network for object goal
  navigation}. In \bibinfo{booktitle}{\emph{International Conference on
  Learning Representations}}.
\newblock


\bibitem[Fang et~al\mbox{.}(2019)]%
        {fang2019scene}
\bibfield{author}{\bibinfo{person}{Kuan Fang}, \bibinfo{person}{Alexander
  Toshev}, \bibinfo{person}{Li Fei-Fei}, {and} \bibinfo{person}{Silvio
  Savarese}.} \bibinfo{year}{2019}\natexlab{}.
\newblock \showarticletitle{Scene memory transformer for embodied agents in
  long-horizon tasks}. In \bibinfo{booktitle}{\emph{Proceedings of the IEEE/CVF
  Conference on Computer Vision and Pattern Recognition}}.
  \bibinfo{pages}{538--547}.
\newblock


\bibitem[Fukushima et~al\mbox{.}(2022)]%
        {fukushima2022object}
\bibfield{author}{\bibinfo{person}{Rui Fukushima}, \bibinfo{person}{Kei Ota},
  \bibinfo{person}{Asako Kanezaki}, \bibinfo{person}{Yoko Sasaki}, {and}
  \bibinfo{person}{Yusuke Yoshiyasu}.} \bibinfo{year}{2022}\natexlab{}.
\newblock \showarticletitle{Object memory transformer for object goal
  navigation}. In \bibinfo{booktitle}{\emph{International Conference on
  Robotics and Automation}}. \bibinfo{pages}{11288--11294}.
\newblock


\bibitem[Gadre et~al\mbox{.}(2022)]%
        {gadre2022continuous}
\bibfield{author}{\bibinfo{person}{Samir~Yitzhak Gadre}, \bibinfo{person}{Kiana
  Ehsani}, \bibinfo{person}{Shuran Song}, {and} \bibinfo{person}{Roozbeh
  Mottaghi}.} \bibinfo{year}{2022}\natexlab{}.
\newblock \showarticletitle{Continuous scene representations for embodied AI}.
  In \bibinfo{booktitle}{\emph{Proceedings of the IEEE/CVF Conference on
  Computer Vision and Pattern Recognition}}. \bibinfo{pages}{14849--14859}.
\newblock


\bibitem[He et~al\mbox{.}(2017)]%
        {he2017mask}
\bibfield{author}{\bibinfo{person}{Kaiming He}, \bibinfo{person}{Georgia
  Gkioxari}, \bibinfo{person}{Piotr Dollar}, {and} \bibinfo{person}{Ross
  Girshick}.} \bibinfo{year}{2017}\natexlab{}.
\newblock \showarticletitle{Mask r-cnn}. In
  \bibinfo{booktitle}{\emph{Proceedings of the IEEE International Conference on
  Computer Vision}}. \bibinfo{pages}{2961--2969}.
\newblock


\bibitem[He et~al\mbox{.}(2016)]%
        {he2016deep}
\bibfield{author}{\bibinfo{person}{Kaiming He}, \bibinfo{person}{Xiangyu
  Zhang}, \bibinfo{person}{Shaoqing Ren}, {and} \bibinfo{person}{Jian Sun}.}
  \bibinfo{year}{2016}\natexlab{}.
\newblock \showarticletitle{Deep residual learning for image recognition}. In
  \bibinfo{booktitle}{\emph{Proceedings of the IEEE Conference on Computer
  Vision and Pattern Recognition}}. \bibinfo{pages}{770--778}.
\newblock


\bibitem[Hochreiter and Schmidhuber(1997)]%
        {hochreiter1997long}
\bibfield{author}{\bibinfo{person}{Sepp Hochreiter} {and}
  \bibinfo{person}{Jurgen Schmidhuber}.} \bibinfo{year}{1997}\natexlab{}.
\newblock \showarticletitle{Long short-term memory}.
\newblock \bibinfo{journal}{\emph{Neural computation}} \bibinfo{volume}{9},
  \bibinfo{number}{8} (\bibinfo{year}{1997}), \bibinfo{pages}{1735--1780}.
\newblock


\bibitem[Hu et~al\mbox{.}(2021)]%
        {hu2021agent}
\bibfield{author}{\bibinfo{person}{Xiaobo Hu}, \bibinfo{person}{Zhihao Wu},
  \bibinfo{person}{Kai Lv}, \bibinfo{person}{Shuo Wang}, {and}
  \bibinfo{person}{Youfang Lin}.} \bibinfo{year}{2021}\natexlab{}.
\newblock \showarticletitle{Agent-centric relation graph for object visual
  navigation}.
\newblock \bibinfo{journal}{\emph{arXiv preprint arXiv:2111.14422}}
  (\bibinfo{year}{2021}).
\newblock


\bibitem[Hu et~al\mbox{.}(2019)]%
        {hu2019acnet}
\bibfield{author}{\bibinfo{person}{Xinxin Hu}, \bibinfo{person}{Kailun Yang},
  \bibinfo{person}{Lei Fei}, {and} \bibinfo{person}{Kaiwei Wang}.}
  \bibinfo{year}{2019}\natexlab{}.
\newblock \showarticletitle{Acnet: Attention based network to exploit
  complementary features for RGBD semantic segmentation}. In
  \bibinfo{booktitle}{\emph{IEEE International Conference on Image
  Processing}}. \bibinfo{pages}{1440--1444}.
\newblock


\bibitem[Izadi et~al\mbox{.}(2011)]%
        {izadi2011kinectfusion}
\bibfield{author}{\bibinfo{person}{Shahram Izadi}, \bibinfo{person}{David Kim},
  \bibinfo{person}{Otmar Hilliges}, \bibinfo{person}{David Molyneaux},
  \bibinfo{person}{Richard Newcombe}, \bibinfo{person}{Pushmeet Kohli},
  \bibinfo{person}{Jamie Shotton}, \bibinfo{person}{Steve Hodges},
  \bibinfo{person}{Dustin Freeman}, \bibinfo{person}{Andrew Davison},
  {et~al\mbox{.}}} \bibinfo{year}{2011}\natexlab{}.
\newblock \showarticletitle{Kinectfusion: real-time 3D reconstruction and
  interaction using a moving depth camera}. In
  \bibinfo{booktitle}{\emph{Proceedings of the 24th Annual ACM Symposium on
  User Interface Software and Technology}}. \bibinfo{pages}{559--568}.
\newblock


\bibitem[Jiang et~al\mbox{.}(2018)]%
        {jiang2018rednet}
\bibfield{author}{\bibinfo{person}{Jindong Jiang}, \bibinfo{person}{Lunan
  Zheng}, \bibinfo{person}{Fei Luo}, {and} \bibinfo{person}{Zhijun Zhang}.}
  \bibinfo{year}{2018}\natexlab{}.
\newblock \showarticletitle{Rednet: Residual encoder-decoder network for indoor
  RGB-D semantic segmentation}.
\newblock \bibinfo{journal}{\emph{arXiv preprint arXiv:1806.01054}}
  (\bibinfo{year}{2018}).
\newblock


\bibitem[Kolve et~al\mbox{.}(2017)]%
        {kolve2017ai2}
\bibfield{author}{\bibinfo{person}{Eric Kolve}, \bibinfo{person}{Roozbeh
  Mottaghi}, \bibinfo{person}{Winson Han}, \bibinfo{person}{Eli VanderBilt},
  \bibinfo{person}{Luca Weihs}, \bibinfo{person}{Alvaro Herrasti},
  \bibinfo{person}{Matt Deitke}, \bibinfo{person}{Kiana Ehsani},
  \bibinfo{person}{Daniel Gordon}, \bibinfo{person}{Yuke Zhu}, {et~al\mbox{.}}}
  \bibinfo{year}{2017}\natexlab{}.
\newblock \showarticletitle{Ai2-thor: An interactive 3D environment for visual
  ai}.
\newblock \bibinfo{journal}{\emph{arXiv preprint arXiv:1712.05474}}
  (\bibinfo{year}{2017}).
\newblock


\bibitem[Li et~al\mbox{.}(2021)]%
        {li2021ion}
\bibfield{author}{\bibinfo{person}{Weijie Li}, \bibinfo{person}{Xinhang Song},
  \bibinfo{person}{Yubing Bai}, \bibinfo{person}{Sixian Zhang}, {and}
  \bibinfo{person}{Shuqiang Jiang}.} \bibinfo{year}{2021}\natexlab{}.
\newblock \showarticletitle{ION: Instance-level object navigation}. In
  \bibinfo{booktitle}{\emph{Proceedings of the 29th ACM International
  Conference on Multimedia}}. \bibinfo{pages}{4343--4352}.
\newblock


\bibitem[Liang et~al\mbox{.}(2021)]%
        {liang2021sscnav}
\bibfield{author}{\bibinfo{person}{Yiqing Liang}, \bibinfo{person}{Boyuan
  Chen}, {and} \bibinfo{person}{Shuran Song}.} \bibinfo{year}{2021}\natexlab{}.
\newblock \showarticletitle{Sscnav: Confidence-aware semantic scene completion
  for visual semantic navigation}. In \bibinfo{booktitle}{\emph{IEEE
  International Conference on Robotics and Automation}}.
  \bibinfo{pages}{13194--13200}.
\newblock


\bibitem[Luo et~al\mbox{.}(2022)]%
        {luo2022stubborn}
\bibfield{author}{\bibinfo{person}{Haokuan Luo}, \bibinfo{person}{Albert Yue},
  \bibinfo{person}{Zhang-Wei Hong}, {and} \bibinfo{person}{Pulkit Agrawal}.}
  \bibinfo{year}{2022}\natexlab{}.
\newblock \showarticletitle{Stubborn: A strong baseline for indoor object
  navigation}. In \bibinfo{booktitle}{\emph{IEEE/RSJ International Conference
  on Intelligent Robots and Systems}}. \bibinfo{pages}{3287--3293}.
\newblock


\bibitem[Maksymets et~al\mbox{.}(2021)]%
        {maksymets2021thda}
\bibfield{author}{\bibinfo{person}{Oleksandr Maksymets},
  \bibinfo{person}{Vincent Cartillier}, \bibinfo{person}{Aaron Gokaslan},
  \bibinfo{person}{Erik Wijmans}, \bibinfo{person}{Wojciech Galuba},
  \bibinfo{person}{Stefan Lee}, {and} \bibinfo{person}{Dhruv Batra}.}
  \bibinfo{year}{2021}\natexlab{}.
\newblock \showarticletitle{Thda: Treasure hunt data augmentation for semantic
  navigation}. In \bibinfo{booktitle}{\emph{Proceedings of the IEEE/CVF
  International Conference on Computer Vision}}. \bibinfo{pages}{15374--15383}.
\newblock


\bibitem[Mayo et~al\mbox{.}(2021)]%
        {mayo2021visual}
\bibfield{author}{\bibinfo{person}{Bar Mayo}, \bibinfo{person}{Tamir Hazan},
  {and} \bibinfo{person}{Ayellet Tal}.} \bibinfo{year}{2021}\natexlab{}.
\newblock \showarticletitle{Visual navigation with spatial attention}. In
  \bibinfo{booktitle}{\emph{Proceedings of the IEEE/CVF Conference on Computer
  Vision and Pattern Recognition}}. \bibinfo{pages}{16898--16907}.
\newblock


\bibitem[Mnih et~al\mbox{.}(2016)]%
        {mnih2016asynchronous}
\bibfield{author}{\bibinfo{person}{Volodymyr Mnih},
  \bibinfo{person}{Adria~Puigdomenech Badia}, \bibinfo{person}{Mehdi Mirza},
  \bibinfo{person}{Alex Graves}, \bibinfo{person}{Timothy Lillicrap},
  \bibinfo{person}{Tim Harley}, \bibinfo{person}{David Silver}, {and}
  \bibinfo{person}{Koray Kavukcuoglu}.} \bibinfo{year}{2016}\natexlab{}.
\newblock \showarticletitle{Asynchronous methods for deep reinforcement
  learning}. In \bibinfo{booktitle}{\emph{International Conference on Machine
  Learning}}. \bibinfo{pages}{1928--1937}.
\newblock


\bibitem[Mnih et~al\mbox{.}(2015)]%
        {mnih2015human}
\bibfield{author}{\bibinfo{person}{Volodymyr Mnih}, \bibinfo{person}{Koray
  Kavukcuoglu}, \bibinfo{person}{David Silver}, \bibinfo{person}{Andrei~A
  Rusu}, \bibinfo{person}{Joel Veness}, \bibinfo{person}{Marc~G Bellemare},
  \bibinfo{person}{Alex Graves}, \bibinfo{person}{Martin Riedmiller},
  \bibinfo{person}{Andreas~K Fidjeland}, \bibinfo{person}{Georg Ostrovski},
  {et~al\mbox{.}}} \bibinfo{year}{2015}\natexlab{}.
\newblock \showarticletitle{Human-level control through deep reinforcement
  learning}.
\newblock \bibinfo{journal}{\emph{nature}} \bibinfo{volume}{518},
  \bibinfo{number}{7540} (\bibinfo{year}{2015}), \bibinfo{pages}{529--533}.
\newblock


\bibitem[Pal et~al\mbox{.}(2021)]%
        {pal2021learning}
\bibfield{author}{\bibinfo{person}{Anwesan Pal}, \bibinfo{person}{Yiding Qiu},
  {and} \bibinfo{person}{Henrik Christensen}.} \bibinfo{year}{2021}\natexlab{}.
\newblock \showarticletitle{Learning hierarchical relationships for object-goal
  navigation}. In \bibinfo{booktitle}{\emph{Conference on Robot Learning}}.
  \bibinfo{pages}{517--528}.
\newblock


\bibitem[Ramakrishnan et~al\mbox{.}(2022)]%
        {ramakrishnan2022poni}
\bibfield{author}{\bibinfo{person}{Santhosh~Kumar Ramakrishnan},
  \bibinfo{person}{Devendra~Singh Chaplot}, \bibinfo{person}{Ziad Al-Halah},
  \bibinfo{person}{Jitendra Malik}, {and} \bibinfo{person}{Kristen Grauman}.}
  \bibinfo{year}{2022}\natexlab{}.
\newblock \showarticletitle{Poni: Potential functions for objectgoal navigation
  with interaction-free learning}. In \bibinfo{booktitle}{\emph{Proceedings of
  the IEEE/CVF Conference on Computer Vision and Pattern Recognition}}.
  \bibinfo{pages}{18890--18900}.
\newblock


\bibitem[Savinov et~al\mbox{.}(2018)]%
        {savinovsemi}
\bibfield{author}{\bibinfo{person}{Nikolay Savinov}, \bibinfo{person}{Alexey
  Dosovitskiy}, {and} \bibinfo{person}{Vladlen Koltun}.}
  \bibinfo{year}{2018}\natexlab{}.
\newblock \showarticletitle{Semi-parametric topological memory for navigation}.
  In \bibinfo{booktitle}{\emph{International Conference on Learning
  Representations}}.
\newblock


\bibitem[Schulman et~al\mbox{.}(2017)]%
        {schulman2017proximal}
\bibfield{author}{\bibinfo{person}{John Schulman}, \bibinfo{person}{Filip
  Wolski}, \bibinfo{person}{Prafulla Dhariwal}, \bibinfo{person}{Alec Radford},
  {and} \bibinfo{person}{Oleg Klimov}.} \bibinfo{year}{2017}\natexlab{}.
\newblock \showarticletitle{Proximal policy optimization algorithms}.
\newblock \bibinfo{journal}{\emph{arXiv preprint arXiv:1707.06347}}
  (\bibinfo{year}{2017}).
\newblock


\bibitem[Sethian(1996)]%
        {sethian1996fast}
\bibfield{author}{\bibinfo{person}{James~A Sethian}.}
  \bibinfo{year}{1996}\natexlab{}.
\newblock \showarticletitle{A fast marching level set method for monotonically
  advancing fronts}.
\newblock \bibinfo{journal}{\emph{Proceedings of the national academy of
  sciences}} \bibinfo{volume}{93}, \bibinfo{number}{4} (\bibinfo{year}{1996}),
  \bibinfo{pages}{1591--1595}.
\newblock


\bibitem[Snavely et~al\mbox{.}(2008)]%
        {snavely2008modeling}
\bibfield{author}{\bibinfo{person}{Noah Snavely}, \bibinfo{person}{Steven~M
  Seitz}, {and} \bibinfo{person}{Richard Szeliski}.}
  \bibinfo{year}{2008}\natexlab{}.
\newblock \showarticletitle{Modeling the world from internet photo
  collections}.
\newblock \bibinfo{journal}{\emph{International journal of computer vision}}
  \bibinfo{volume}{80} (\bibinfo{year}{2008}), \bibinfo{pages}{189--210}.
\newblock


\bibitem[Vaswani et~al\mbox{.}(2017)]%
        {vaswani2017attention}
\bibfield{author}{\bibinfo{person}{Ashish Vaswani}, \bibinfo{person}{Noam
  Shazeer}, \bibinfo{person}{Niki Parmar}, \bibinfo{person}{Jakob Uszkoreit},
  \bibinfo{person}{Llion Jones}, \bibinfo{person}{Aidan~N Gomez},
  \bibinfo{person}{Lukasz Kaiser}, {and} \bibinfo{person}{Illia Polosukhin}.}
  \bibinfo{year}{2017}\natexlab{}.
\newblock \showarticletitle{Attention is all you need}.
\newblock \bibinfo{journal}{\emph{Advances in neural information processing
  systems}}  \bibinfo{volume}{30} (\bibinfo{year}{2017}).
\newblock


\bibitem[Wortsman et~al\mbox{.}(2019)]%
        {worts}
\bibfield{author}{\bibinfo{person}{Mitchell Wortsman}, \bibinfo{person}{Kiana
  Ehsani}, \bibinfo{person}{Mohammad Rastegari}, \bibinfo{person}{Ali Farhadi},
  {and} \bibinfo{person}{Roozbeh Mottaghi}.} \bibinfo{year}{2019}\natexlab{}.
\newblock \showarticletitle{Learning to learn how to learn: Self-adaptive
  visual navigation using meta-learning}. In
  \bibinfo{booktitle}{\emph{Proceedings of the IEEE/CVF Conference on Computer
  Vision and Pattern Recognition}}. \bibinfo{pages}{6750--6759}.
\newblock


\bibitem[Wu et~al\mbox{.}(2019)]%
        {wu2019bayesian}
\bibfield{author}{\bibinfo{person}{Yi Wu}, \bibinfo{person}{Yuxin Wu},
  \bibinfo{person}{Aviv Tamar}, \bibinfo{person}{Stuart Russell},
  \bibinfo{person}{Georgia Gkioxari}, {and} \bibinfo{person}{Yuandong Tian}.}
  \bibinfo{year}{2019}\natexlab{}.
\newblock \showarticletitle{Bayesian relational memory for semantic visual
  navigation}. In \bibinfo{booktitle}{\emph{Proceedings of the IEEE/CVF
  International Conference on Computer Vision}}. \bibinfo{pages}{2769--2779}.
\newblock


\bibitem[Yang et~al\mbox{.}(2018)]%
        {yang2018visual}
\bibfield{author}{\bibinfo{person}{Wei Yang}, \bibinfo{person}{Xiaolong Wang},
  \bibinfo{person}{Ali Farhadi}, \bibinfo{person}{Abhinav Gupta}, {and}
  \bibinfo{person}{Roozbeh Mottaghi}.} \bibinfo{year}{2018}\natexlab{}.
\newblock \showarticletitle{Visual semantic navigation using scene priors}.
\newblock \bibinfo{journal}{\emph{arXiv preprint arXiv:1810.06543}}
  (\bibinfo{year}{2018}).
\newblock


\bibitem[Ye and Yang(2021)]%
        {ye2021hierarchical}
\bibfield{author}{\bibinfo{person}{Xin Ye} {and} \bibinfo{person}{Yezhou
  Yang}.} \bibinfo{year}{2021}\natexlab{}.
\newblock \showarticletitle{Hierarchical and partially observable goal-driven
  policy learning with goals relational graph}. In
  \bibinfo{booktitle}{\emph{Proceedings of the IEEE/CVF Conference on Computer
  Vision and Pattern Recognition}}. \bibinfo{pages}{14101--14110}.
\newblock


\bibitem[Zhang et~al\mbox{.}(2022)]%
        {zhang2022generative}
\bibfield{author}{\bibinfo{person}{Sixian Zhang}, \bibinfo{person}{Weijie Li},
  \bibinfo{person}{Xinhang Song}, \bibinfo{person}{Yubing Bai}, {and}
  \bibinfo{person}{Shuqiang Jiang}.} \bibinfo{year}{2022}\natexlab{}.
\newblock \showarticletitle{Generative meta-adversarial network for unseen
  object navigation}. In \bibinfo{booktitle}{\emph{European Conference on
  Computer Vision}}. \bibinfo{pages}{301--320}.
\newblock


\bibitem[Zhang et~al\mbox{.}(2021)]%
        {zhang2021hi}
\bibfield{author}{\bibinfo{person}{Sixian Zhang}, \bibinfo{person}{Xinhang
  Song}, \bibinfo{person}{Yubing Bai}, \bibinfo{person}{Weijie Li},
  \bibinfo{person}{Yakui Chu}, {and} \bibinfo{person}{Shuqiang Jiang}.}
  \bibinfo{year}{2021}\natexlab{}.
\newblock \showarticletitle{Hierarchical object-to-zone graph for object
  navigation}. In \bibinfo{booktitle}{\emph{Proceedings of the IEEE/CVF
  International Conference on Computer Vision}}. \bibinfo{pages}{15130--15140}.
\newblock


\bibitem[Zhu et~al\mbox{.}(2019)]%
        {zhuepisodic}
\bibfield{author}{\bibinfo{person}{Guangxiang Zhu}, \bibinfo{person}{Zichuan
  Lin}, \bibinfo{person}{Guangwen Yang}, {and} \bibinfo{person}{Chongjie
  Zhang}.} \bibinfo{year}{2019}\natexlab{}.
\newblock \showarticletitle{Episodic reinforcement learning with associative
  memory}. In \bibinfo{booktitle}{\emph{International Conference on Learning
  Representations}}.
\newblock


\bibitem[Zhu et~al\mbox{.}(2017)]%
        {zhu2017target}
\bibfield{author}{\bibinfo{person}{Yuke Zhu}, \bibinfo{person}{Roozbeh
  Mottaghi}, \bibinfo{person}{Eric Kolve}, \bibinfo{person}{Joseph~J Lim},
  \bibinfo{person}{Abhinav Gupta}, \bibinfo{person}{Li Fei-Fei}, {and}
  \bibinfo{person}{Ali Farhadi}.} \bibinfo{year}{2017}\natexlab{}.
\newblock \showarticletitle{Target-driven visual navigation in indoor scenes
  using deep reinforcement learning}. In \bibinfo{booktitle}{\emph{IEEE
  International Conference on Robotics and Automation}}.
  \bibinfo{pages}{3357--3364}.
\newblock


\end{thebibliography}










\end{document}